\documentclass[lettersize,journal]{IEEEtran}
\usepackage{amsmath,amsfonts}
\usepackage{algorithmic}
\usepackage{algorithm}
\usepackage{array}
\usepackage[caption=false,font=normalsize,labelfont=sf,textfont=sf]{subfig}
\usepackage{textcomp}
\usepackage{stfloats}
\usepackage{url}
\usepackage{verbatim}
\usepackage{graphicx}
\usepackage{cite}
\usepackage{xcolor}
\usepackage{enumitem}

\newcommand{\T}{{\top}}

\DeclareMathOperator{\sigmoid}{sigmoid}

\setlist[itemize]{leftmargin=*}

\begin{document}

\title{A Discriminative Hierarchical PLDA-based Model \\ for Spoken Language Recognition}

\author{Luciana Ferrer, Diego Castan, Mitchell McLaren, Aaron Lawson
\thanks{Ferrer is with Instituto de Investigaci\'{o}n en Ciencias de la Computaci\'{o}n (ICC), CONICET-UBA, Argentina. Castan, McLaren and Lawson are with SRI International, USA.}
\thanks{Manuscript received April 19, 2021; revised August 16, 2021.}}

\markboth{Journal of \LaTeX\ Class Files,~Vol.~14, No.~8, August~2021}%
{Shell \MakeLowercase{\textit{et al.}}: A Sample Article Using IEEEtran.cls for IEEE Journals}


\maketitle

\begin{abstract}
Spoken language recognition (SLR) refers to the automatic process used to determine the language present in a speech sample. SLR is an important task in its own right, for example, as a tool to analyze or categorize large amounts of multi-lingual data. Further, it is also an essential tool for selecting downstream applications in a work flow, for example, to chose appropriate speech recognition or machine translation models. SLR systems are usually composed of two stages, one where an embedding representing the audio sample is extracted and a second one which computes the final scores for each language. In this work, we approach the SLR task as a detection problem and implement the second stage as a probabilistic linear discriminant analysis (PLDA) model. We show that discriminative training of the PLDA parameters gives large gains with respect to the usual generative training. Further, we propose a novel hierarchical approach where two PLDA models are trained, one to generate scores for clusters of highly-related languages and a second one to generate scores conditional to each cluster. The final language detection scores are computed as a combination of these two sets of scores. The complete model is trained discriminatively to optimize a cross-entropy objective. We show that this hierarchical approach consistently outperforms the non-hierarchical one for detection of highly related languages, in many cases by large margins. We train our systems on a collection of datasets including over 100 languages, and test them both on matched and mismatched conditions, showing that the gains are robust to condition mismatch.
\end{abstract}


\section{Introduction}
\label{sec:intro}

\IEEEPARstart{T}{he} problem of recognizing the language present in a speech signal has been of interest in the speech processing community for decades \cite{martin2014nist,li2013spoken}. Some of the most standard and successful approaches for language recognition consist of two or more stages. First, a deep neural network (DNN) model is trained to predict phone categories (usually senones) for one or more languages. This DNN includes a relatively low-dimensional layer, called bottleneck (BN), which is then used a feature extractor \cite{jiang:2014,matejka:2014,Ferrer:aslp15,mclaren:icassp16}. The BN features can then be used as input features to extract i-vectors \cite{Dehak11} for each sample. Finally, a backend is used to produce the  language scores. Alternatively, instead of computing i-vectors, many systems use the BN features as input to another DNN trained to classify the languages available in the training data \cite{snyder2018spoken}. This language DNN can be used to extract embeddings given by the output of some internal layer in the network that can then be used, like the i-vectors, as the input to a backend that produces the final language scores. When the test data is assumed to contain a subset of the training languages, the language DNN can be used directly to perform classification \cite{cai2019utterance,villalba2018end}. 

In practice, a common scenario is for the set of test languages to be undefined during development of the system. In this case, the system has to be capable of labelling a sample as an unknown language if it does not belong to any of the languages available in the system. Further, in many scenarios the user of the system might prefer to see all plausible languages as candidates rather than just the best one. For these reasons, the problem of language recognition is often posed as a detection problem, where each sample can be labelled with zero, one or more language labels. This is the scenario used in NIST language recognition evaluations \cite{martin2003nist,Martin:2009,zhao2016results,sadjadi20182017} and, more recently, in the Oriental Language Recognition challenges \cite{wang2016ap16,li2021oriental}, and is the scenario we consider in this paper.

Probabilistic linear discriminant analysis (PLDA) is a standard method to generate detection scores for face and speaker verification \cite{Ioffe:2006,prince:plda}, and has also been successfully used for language detection \cite{mccree2014multiclass,cumani2015exploiting}. PLDA provides a way to compute likelihood ratios (LR) scores for  each of the detectors, under Gaussian assumptions. While there are several flavors of PLDA, the most commonly used one both for speaker verification and for language detection is the two-covariance model \cite{Brummer:odyssey10,sizov2014}. Either i-vectors or embeddings are used as input to PLDA. Another backend commonly used for language detection is the Gaussian linear classifier \cite{martinez:2011}. In this case, though, the model can only be used to compute likelihoods, not LRs and, hence, it can only be used in closed-set scenarios \cite{cumani2015exploiting,mccree2014multiclass}. 

The two-covariance PLDA model assumes that the vector representing a signal has a Gaussian distribution around a language-dependent mean. Further, it assumes that the language-dependent means are, in turn, also Gaussian distributed. Yet, in most language recognition datasets, the set of available languages is unlikely to be well modelled by this Gaussian assumption, since some languages come in highly-related clusters, while others have little in common to any other language in the dataset. To address this concern, in this work, we explore an extension of the PLDA method where the Gaussian assumption is applied to clusters of languages rather than to individual languages. This model is used to generate per-cluster LRs. Then, a second PLDA stage generates by-language LRs, conditional to the cluster, which are then combined with the per-cluster LRs to generate the final LR. The full model is trained discriminatively, as explored for the standard PLDA model in several previous papers \cite{burget:icassp11,mccree2014multiclass,snyder2016deep,Rohdin2018,magnet:odyssey2020}, including recent works from our group \cite{ferrer2020discriminative,ferrer2020speaker,ferrer2022speaker}.

Our goal is to develop a general purpose language recognition system, capable of detecting a large number of languages. To this end, we train our systems on a collection of datasets including a total of 100 languages, using various augmentation techniques. We evaluate our systems on a number of different test sets, including NIST LRE data, BABEL, KALAKA, and many others. We show (1) that discriminative training of the standard PLDA model gives a large improvement over generative training on most datasets, and (2) that our proposed hierarchical model leads to further improvements, especially for closely related languages. The code used to run the experiments in this paper is available at \url{https://github.com/luferrer/DCA-PLDA}.

\section{PLDA Formulation}

In this work, we assume that an embedding extractor is available to convert each audio sample into a single vector of fixed dimension $D$. These vectors are first processed with linear discriminant analysis (LDA) to reduce their dimension, and the resulting vectors are normalized to have zero mean and unit variance in each component and, finally, length normalized \cite{romero:lennorm}. Language detection scores are then generated using a two-covariance PLDA model~\cite{Brummer:odyssey10} which assumes that each embedding, $w$, can be modeled as
\begin{equation*}
w = y + e,
\end{equation*}
where $y$ and $e$ are vectors of dimension $D$, are assumed to be independent, and are both Gaussian-distributed so that
\begin{eqnarray}
y & \sim & \mathcal{N}(\mu,B^{-1}), \label{eq:y}\\
w|y & \sim & \mathcal{N}(y, W^{-1}),
\end{eqnarray}
where $B$ is the between-class precision matrix and $W$ is the within-class precision matrix, both with dimension $D\times D$. The variable $y$ depends on the language and determines the mean of the embedding distribution for that language. 

This formulation can then be used to compute language detection scores. In this scenario, a set of $S$ enrollment samples $\Omega_l=\{w_{l1},\ldots,w_{lS}\}$ is available for each language  of interest,~$l$. Given a test sample $w$, the detection score for each language is computed as the following log-likelihood ratio (LLR)
\begin{eqnarray}
L_{l} & = & \log \frac{P(w, \Omega_l|H_s)}{P(w, \Omega_l|H_d)} \label{eq:llr}
\end{eqnarray}
where $H_s$ and $H_d$ are the hypotheses that the speech in $w$ is spoken in the same or different language as the speech in $\Omega_l$, respectively. This LLR can be computed in closed-form as a function of $B$, $W$, $\mu$, the mean of the $\Omega_l$ vectors, $S$, and the test vector $w$. This formula can be found in Equation (23) in \cite{Brummer:odyssey10}. We call this the ``exact'' scoring formula, and we use it to compute the scores for the PLDA approach.

In this work, we compare the performance of the PLDA backend where the parameters $B$, $W$ and $\mu$ are trained in the traditional way, using the Expectation-Maximization algorithm \cite{em4splda}, with the performance of a discriminatively trained version of this same backend. A discriminative approach for training a Gaussian classifier for open-set SLR was proposed in \cite{mccree2014multiclass}, where the parameters of the model are updated using Maximum Mutual Information. In our case, as explained next, we optimize a different set of parameters which are the ones needed during scoring, and training is done using a stochastic gradient descent approach. 

When a single enrollment vector, $w_l$, is available, the LLR for test sample $w$ has the following simple form:
\begin{eqnarray}
L_l = 2 w ^\T \Lambda w_l + w^\T \Gamma w + w_l^\T \Gamma w_l + w^\T c + w_l^\T c + k,  \label{eq:plda_score}
\end{eqnarray}
where $\Lambda$, $\Gamma$, $c$, and $k$ are functions of $B$, $W$ and $\mu$ (see, for example, \cite{cumani2013pairwise,glembek:11}). In our previous work, we proposed a generalization of the PLDA backend where we introduce a condition-dependent calibration stage and train the parameters of the whole model discriminatively \cite{ferrer2020speaker,ferrer2020discriminative,ferrer2022speaker}. In that work we focused on the speaker verification problem and used the simplified scoring formulation above, discriminatively training the $\Lambda$, $\Gamma$, $c$ and $k$ parameters directly, rather than $B$, $W$ and $\mu$. This allows us to avoid the matrix inversions that are needed to compute the first set of parameters from the second set, but has the disadvantage of not allowing us to use the exact scoring formula when more than one enrollment sample is available, as is the case in language detection. 

In speaker verification, a common approach for using Equation (\ref{eq:plda_score}) when more than one enrollment sample is available is to replace the available samples with their mean and then plug it in the scoring formula as a single enrollment vector. In this work, we adopt this approximation when training the model discriminatively since, as described in Section \ref{sec:configuration}, preliminary experiments using the exact scoring formulation for discriminative training did not result in improved performance.
Hence, in the approach we call discriminative PLDA (DPLDA), we discriminatively train $\Lambda$, $\Gamma$, $c$, $k$, as well as the vectors $w_l$ that represent each language, and the LDA parameters. The LDA parameters are initialized with the  standard maximum-likelihood estimates. The parameters $\Lambda$, $\Gamma$, $c$, $k$ are initialized using the $B$, $W$ and $\mu$ obtained with EM training and the formulas described in \cite{cumani2013pairwise,glembek:11} to convert one set of parameters to the other. The vectors $w_l$ are initialized with the mean of all the samples for each language $l$. Hence, at initialization, the DPLDA model outputs scores that are identical to PLDA when the approximate scoring formulation is used. The training criterion and procedure are explained in more detail in the next section, since it is the same procedure used to train the proposed hierarchical approach. 

\section{Hierarchical PLDA Formulation}
\label{sec:hier}

\begin{figure}
\includegraphics[width=\linewidth]{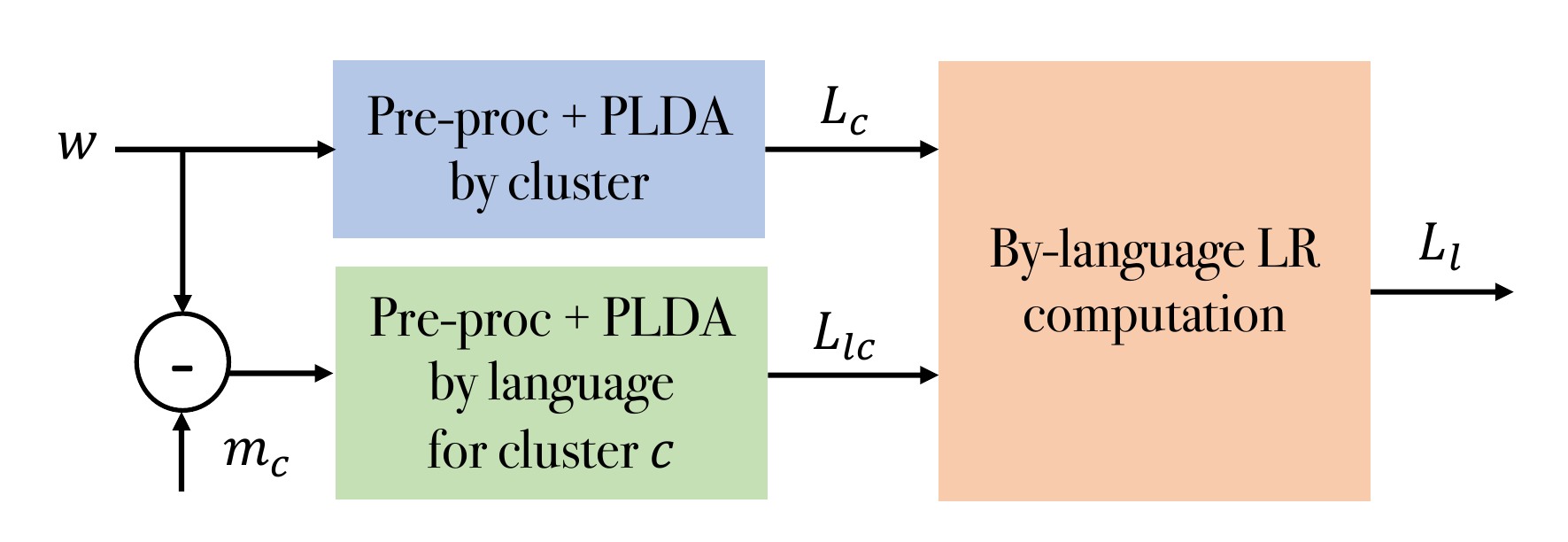}
\caption{Proposed hierarchical model. The figure shows how the LLR for language $l$, which belongs to cluster $c$, is computed for a certain input embedding $w$. The embedding $w$ is processed by a first PLDA stage that generates by-cluster LLRs, $L_c$. In parallel, a vector $m_c$ for cluster $c$ is subtracted from $w$. The resulting vector is then processed by a second PLDA stage that generates cluster-conditional LLRs, $L_{lc}$. Finally, $L_l$ is computed as a function of $L_c$ and $L_{lc}$ using Equation (\ref{eq:Ll}).
}
\label{fig:method}
\end{figure}

The $y$ in Equation (\ref{eq:y}) is assumed to be Gaussian-distributed, which, as mentioned in the introduction, is an assumption that is unlikely to hold well for most language recognition datasets, since some languages usually group into tight clusters while others are not highly related to any other language in the set. To address this issue, we propose a hierarchical model composed of two levels. In a pre-processing stage, all training languages are clustered based on their mean vector. Then, the first level of the hierarchical approach consists of a PLDA model (including the LDA with mean and variance normalization and length-normalization stages) which generates LLRs for each of the clusters. In this model, each cluster detector is represented by a vector, of dimension equal to the LDA dimension in that stage, which is used as the $w_l$ in Equation (\ref{eq:plda_score}). In the second level, for each cluster, a vector $m_c$ representing each cluster is subtracted from the input vector before estimating a new LLR for each detector, conditional to the cluster. In this level, each language is represented by a vector $w_l$, as in the flat approach, except that, in this case, the vector is encouraged to be relative to the cluster since the inputs to this level are given by the embeddings shifted by a cluster-dependent vector $m_c$. Finally, the two LLRs are combined as described next, to compute the final per-language LLR. Figure \ref{fig:method} shows a schematic of the proposed approach.

\subsection{Score Computation}

The model needs to generate LLRs for every language. As explained next, we compute these LLRs as a function of the LLR for each cluster and the cluster-conditional LLR.  
In this section, we find it convenient to work in terms of likelihood-ratios (LRs), which we will call $\Lambda$, instead of LLRs. 

For language $l$ and test vector $w$, we want the model to output (the logarithm of)
\begin{eqnarray}
\Lambda_l = e^{L_l} = \frac{p(w|l)}{p(w|\hat l)} \label{eq:lambdal1}
\end{eqnarray}
where we have simplified the notation of Equation (\ref{eq:llr}) by making $\Omega_l$ implicit and calling the two hypothesis $l$ and $\hat l$ (not $l$). Assuming that language  $l$ belongs to cluster $c$, and using Bayes rule, we obtain
\begin{eqnarray*}
p(w|l) \hspace{-0.22cm}& = & \hspace{-0.22cm}p(w|l,c) = \frac{p(l|c,w) p(c|w) p(w)}{p(l|c)p(c)} \\
p(w|\hat l) \hspace{-0.22cm}& = & \hspace{-0.22cm}p(w|(\hat l, c) \lor \hat c)
 =  \frac{\left(p(\hat l|c,w) p(c|w) + p(\hat c | w) \right)p(w)}{p(\hat l|c) p(c) + p(\hat c)} 
\end{eqnarray*}
where $\hat c$ refers to any cluster that is not cluster $c$.
Plugging those expressions in Equation (\ref{eq:lambdal1}) we get
\begin{eqnarray}
\Lambda_l  =  \frac{p(l|c,w) p(c|w) }{p(\hat l|c,w)p(c|w) + p(\hat c | w)}\frac{p(\hat l|c) p(c) + p(\hat c)}{p(l|c)p(c)} 
 = \frac{O_l}{P_l}
 \label{eq:lambdal}
\end{eqnarray}
where $O_l$ and $P_l$ refer to the posterior and prior odds, given by the first factor and the inverse of the second factor, respectively.
The first PLDA model in our proposed approach, generates the LRs for each cluster $c$,
\begin{eqnarray}
\Lambda_c = e^{L_c} = \frac{p(w|c)}{p(w|\hat c)} =  \frac{p(c|w)}{p(\hat c|w)}\frac{p(\hat c)}{p(c)} = \frac{O_c}{P_c}
\label{eq:lambdac}
\end{eqnarray}
while the second PLDA model generates the LR for language $l$ given that we know the cluster is $c$:
\begin{eqnarray}
\Lambda_{lc} = e^{L_{lc}} = \frac{p(w|l,c)}{p(w|\hat l,c)} =  \frac{p(l|c,w)}{p(\hat l|c,w)}\frac{p(\hat l| c)}{p(l|c)} = \frac{O_{lc}}{P_{lc}}
\label{eq:lambdalc}
\end{eqnarray}
for any $l$ that belongs to cluster $c$, and is zero otherwise.

In these equations, the $O$'s and $P$'s are the corresponding  posterior and prior odds, from which the posteriors and priors can be recovered as follows:
\begin{align}
p(l|c,w) & = \frac{O_{lc}}{1+O_{lc}}, &  p(c|w) & = \frac{O_{c}}{1+O_{c}} \\
p(l|c) & = \frac{P_{lc}}{1+P_{lc}}, &  p(c) & = \frac{P_{c}}{1+P_{c}} \label{eq:priors}
\end{align}

Replacing these values in Equation \ref{eq:lambdal} and taking the logarithm, we get:
\begin{eqnarray}
L_l = \log \Lambda_l   = \log \left(
\frac{O_c O_{lc} }{O_c + O_{lc} + 1} \frac{P_c + P_{lc}+1}{P_c P_{lc}}\right) \label{eq:Ll}
\end{eqnarray}
where
\begin{align}
O_c   & =  e^{L_{c}} P_c,        & O_{lc} & =  e^{L_{lc}} P_{lc},\\
P_c   & = \frac{p(c)}{1-p(c)}, & P_{lc} & =  \frac{p(l|c)}{1-p(l|c)}.
\end{align}
The posterior odds ($O_c$ and $O_{lc}$) are obtained from Equations (\ref{eq:lambdac}) and (\ref{eq:lambdalc}) and are a function of the output of the two PLDA models, $L_c$ and $L_{lc}$. The prior odds ($P_c$ and $P_{lc}$) are obtained from Equations (\ref{eq:priors}) and are a function of the priors for each cluster $p(c)$ and each language given the cluster $p(l|c)$. 
In our work, we take the prior for each cluster $p(c)$ to be equal to the number of languages in that cluster divided by the total number of languages, and $p(l|c)$ to be equal to the inverse of the number of languages in that cluster.

Note that, given a certain sample, the LLR for language $l$ corresponding to cluster $c$ is computed using the output corresponding to cluster $c$ of the by-cluster PLDA model ($L_c$) and the output corresponding to language $l$ of the by-language PLDA model for cluster $c$ ($L_{lc}$). The true cluster of the sample is not needed for this computation; only the assignment of languages to clusters is required. This assignment is obtained with a clustering stage (explained in Section \ref{sec:clusters}) which is done once, before model training. After that, the assignment is fixed throughout training and testing. 

In this work, the parameters of the by-language cluster-conditional PLDA model are shared across clusters. This stage takes as input the embeddings after subtracting a cluster-dependent vector (see Figure \ref{fig:method}). After this subtraction, the model assumes that the cluster-conditional LR can be computed using the same PLDA parameters for all clusters. While this is probably suboptimal for some clusters, where the languages might differ from each other in ways that are different from those in other clusters, our results indicate that the assumption works very well in practice.

\subsection{Model training}

The full model is trained discriminatively with stochastic gradient descent using a dataset of embeddings labelled with language class, and the table of correspondence between language and cluster.  As done in \cite{ferrer2022speaker}, before discriminative training, the parameters of the model are initialized with reasonable estimates. The full initialization process is as follows:
\begin{itemize}
\item Initialize $m_c$ as the average of the per-language average embeddings for cluster $c$. That is, at initialization, $m_c = \frac{1}{|{\cal C}_c|}\sum_{l\in {\cal C}_c} m_l$, where ${\cal C}_c$ is the set of languages in cluster  $c$, $|{\cal C}_c|$ is the number of languages in that cluster, and $m_l$ is the average embedding for language $l$.
\item Initialize the by-cluster PLDA model with the maximum likelihood estimates of the parameters using the training embeddings and the clusters as class labels.
\item Initialize the by-language PLDA model with the maximum likelihood estimates of the parameters using the training embeddings after subtracting the $m_c$ corresponding to each sample, and the language as class label. 
\end{itemize}
Note that the last step is just a heuristic to get a reasonable starting point for the by-language PLDA, but it is strictly not correct since the maximum likelihood estimates do not take into account the hierarchical nature of the model (i.e., the fact that the output of this second stage is meant to be a conditional LR rather than the actual LR). Nevertheless, this heuristic procedure for initialization gives a reasonable starting point for  discriminative training. 

After initialization, stochastic gradient descent is used to fine-tune the parameters by optimizing a binary cross-entropy loss. To compute this loss over a training mini-batch of size $B$, each sample in the batch is tested against every language detector, creating \emph{trials} which are either positive, if the sample's language coincides with that of the detector, or negative otherwise. The loss is computed as a weighted average over all trials in the batch, weighting the positive and the negative samples separately with a tunable weight $\pi$:
\begin{equation}
    \mathcal{L}_{\text{lan}} = - \frac{\pi}{P} \sum_l  \sum_{i | l_i=l}  \log(q_{l}^{(i)}) - \frac{1-\pi}{N} \sum_l  \sum_{i | l_i \neq l}  \log(1-q_{l}^{(i)})  \label{eq:crossent}
\end{equation}
where 
\begin{equation}
q_{l}^{(i)} = \sigmoid \left(L_{l}^{(i)} + \log \frac{\pi}{1-\pi} \right)
\end{equation}
and where $i$ is the sample index which runs from 1 to $B$, $l$ is the language index which runs from 1 to $L$, the total number of detectors in the model, $l_i$ is the language for sample $i$, $L_{l}^{(i)}$ is the system's output for sample $i$ for language detector $l$, and $P = \sum_l \sum_{i | l_i=l} 1$, and $N = \sum_l \sum_{i | l_i \neq l} 1$. $N$ and $P$ are the number of negative and positive trials, given by $B*(L-1)$ and $B$, respectively, since every sample in the batch generates one positive trial and $L-1$ negative trials.

When training the hierarchical model, we can also compute a binary cross-entropy for the task of detecting the cluster of each sample, rather than the language, $\mathcal{L}_{\text{clu}}$. The final loss can then be computed as:
\begin{equation}
    \mathcal{L} = (1-\alpha)\ \mathcal{L}_{\text{lan}} + \alpha\ \mathcal{L}_{\text{clu}}
\end{equation}
where $\alpha$ is a tunable parameter and $\mathcal{L}_{\text{clu}}$ is computed analogously to $\mathcal{L}_{\text{lan}}$ but using the cluster labels instead of the language labels.

\section{Experimental Setup}

In this section we describe the training, development and evaluation data, the embedding extractor, and the system configuration used for the experiments.

\subsection{Datasets}

The training data for the backend is given by a collection of different datasets: the data used at SRI for the 2009 NIST language recognition evaluation (LRETR), RATS data, NIST speaker recognition evaluation data (SRE), Voxlingua107 (VOXL), Panarabic (PANA), Callhome (CALLH) and Callfriend (CALLF). A description of these datasets, including references, data curation procedure, number of languages and number of samples, can be found in Appendix \ref{app:train_data}. In total, the resulting dataset includes 100 languages and a total of 248 thousand waveforms.  The number of samples per language varies widely from a few dozen to tens of thousands. A small percentage of data from each of the sets is held-out for testing. The training data is further processed to include 4 degraded files for each original file: one with added music, one with added noise, one with reverb and one transcoded. A description of the types of augmentations used can be found in \cite{ferrer2022speaker}. Finally, each resulting file, including the raw one, is chunked into 4 segments with speech duration between 3 and 30 seconds. The chunk duration is randomly selected to result in a uniform distribution in the logarithmic domain. These chunks are then used for training the backends. 

For development, hyperparameter tuning and selection of the best seed and epoch, we use the 7 sets held-out from the training sets as matched conditions, and the BABEL dataset as mismatched condition. 
The evaluation data, which was not used for development, is composed of the following datasets: LRE15, LRE17, LASRS, KALAKA and Crowdsource (CROWD). A detailed description of these development and evaluation datasets and their statistics can be found in Appendix \ref{app:eval_data}. Note that, given the large number of languages available in our training data, very few of the test languages do not have a corresponding detector and, as a consequence, are out-of-set languages. Table \ref{table:stats} shows the percentage of samples in each dataset that are out-of-set.

Each of the development and evaluation samples are cut to create new chunks that contain 4, 8, 16 and 32 seconds of speech to be able to test performance as a function of speech duration. To do this, for each sample we find the most speech-dense region containing approximately the desired amount of speech, and at least 66\% speech activity in the region. The motivation here is to avoid back-channel speech (``um'', ``uh-huh'') and rather capture content-rich speech. Finally, when a sample is shorter than the target duration, we include it without chunking if the duration is larger than half the target duration, otherwise, we discard it.
All datasets have a majority of samples longer than 32 seconds and, hence, we are able to create approximately the same amount of chunks for all target durations.  

The language labels in the different training and test datasets were mapped to ISO-639-3 to use a common convention. Details on the labelling process can be found in Appendix \ref{app:labelling}. The lists and labels used in our experiments for the different datasets can be provided upon request.

\subsection{Embedding Extractor}
\label{sec:embeddings}
As input to our embedding extractor, we use 80-dimensional bottleneck features extracted using a DNN trained on English data to classify English senones. These features have been shown to work better than traditional low-level features like MFCCs for the task of language recognition \cite{Ferrer:aslp15,snyder18_odyssey}. 
The network is trained to classify 3450 senones (tri-phone states) using Fisher~\cite{cieri:04} and Switchboard~\cite{godfrey:92} data, and has 5 hidden layers of 600 nodes with the last hidden layer being the bottleneck of 80 nodes. The input to this DNN is given by power-normalized cepstral coefficients (PNCC)~\cite{kim:12} features. A DNN-based speech activity detection (SAD) system is then used to filter out non-speech frames. Details on the SAD system can be found in Appendix A in \cite{ferrer2022speaker}.

The bottleneck features over speech frames are fed into a standard x-vector architecture \cite{snyder2016deep,KaldiRecipe17}, except that we use a modified context for the initial layers  and slightly smaller layer sizes compared to the standard configuration. Specifically, instead of using a [-2,-1,0,1,2] time-delay indices for the first layer, [-1,0,1] for the second layer, and [-1,0,1] for the third layer, we use [-10,-5,0,5,10] for the first layer, [-10,0,10] for the second layer, and [-10,0,10] for the third layer. In total, the DNN has 7 hidden layers with sizes 512, 384, 384, 384, 1250, 384, and 384. The output of the 6th layer in this DNN is used as embedding extractor, so  embeddings are 384-dimensional.
The model is trained with the procedure described in \cite{ferrer2022speaker} (Appendix A), using the LRETR, RATS, SRE and VOXL datasets described in \ref{app:train_data} with chunks of 8 seconds over 3 epochs.

While some improvements in performance could probably be achieved with more recent architectures for the embedding extractor, or using senone bottleneck features extracted from a DNN trained with multiple languages, we believe the conclusions in this paper would still hold with those improved embedding extractors. Testing this hypothesis is left for future work. 

\subsection{Language Clusters}
\label{sec:clusters}

\begin{figure*}
\centering
\includegraphics[height=0.80\textheight]{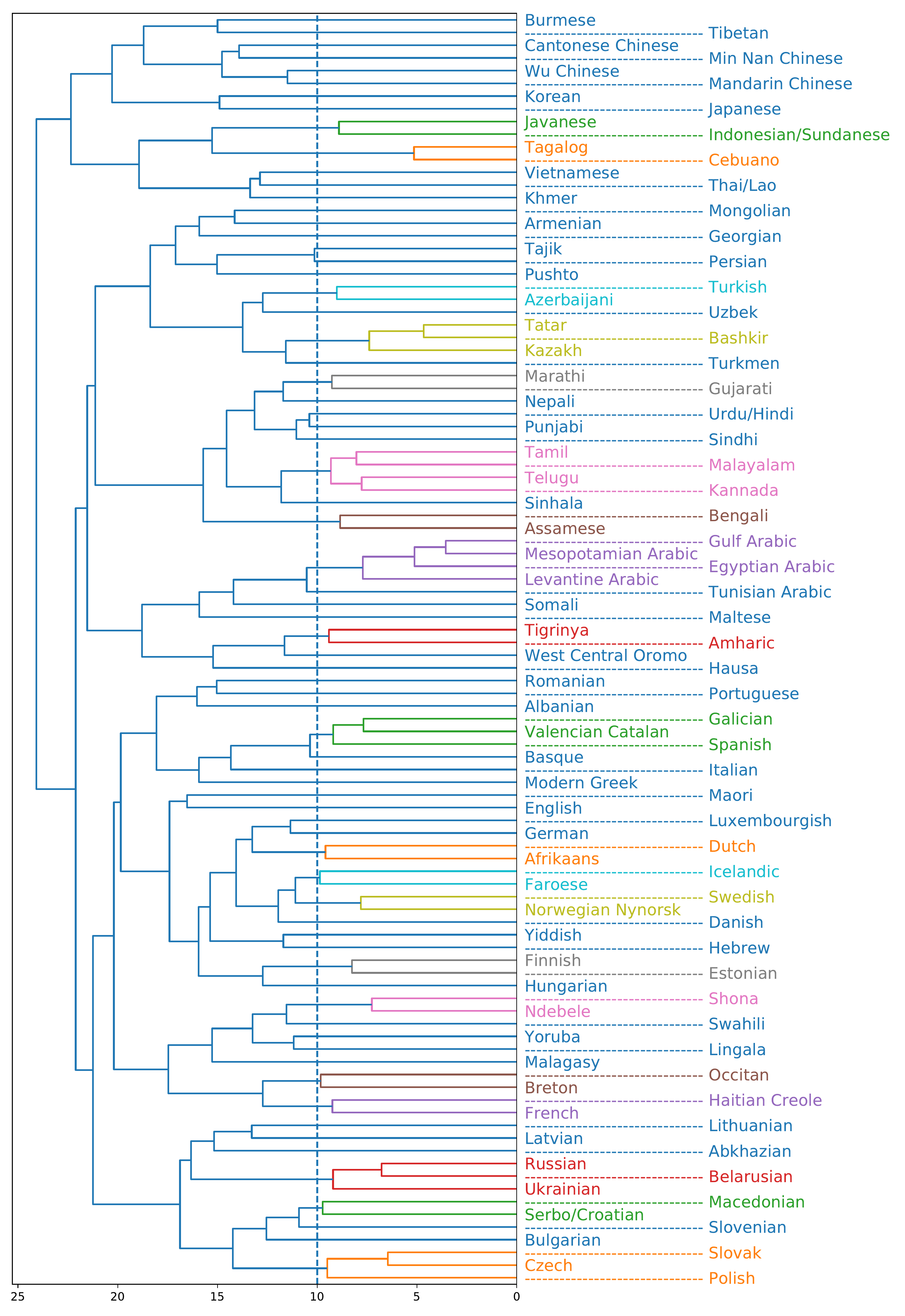}
\caption{Language dendogram. The colors of the lines and language names indicate the clusters. Blue indicates that the language is not clustered with any other. The dashed vertical line indicates the distance threshold used for clustering. Any two languages with a distance smaller than this threshold are clustered together in the hierarchical model.}
\label{fig:clusters}
\end{figure*}

The language clusters were determined by clustering the mean embedding for each of the languages available in the training data. The mean embedding for each language was obtained using one chunk for each original file (files with added degradation are not used for this purpose) with approximately 30 seconds of speech. The resulting mean embeddings were then clustered using agglomerative clustering with average linkage method and a distance matrix given by a PLDA model trained on the training data described in Appendix~\ref{app:train_data}. The distance threshold, which determines the number of clusters, was tuned to optimize the average performance on the development sets. 
Figure \ref{fig:clusters} shows the clusters we obtained with this procedure. The plot was created using the dendogram method in scipy.cluster.hierarchy. 
With the optimized distance threshold of 10, we obtain 20 clusters with two or more languages. Languages that are not clustered with any other language are considered each as a separate cluster. Overall, we have a total of 72 clusters, 52 of them single-language.

A reasonable question to ask is whether the mean vectors for a certain language, computed separately for each dataset, would cluster with each other. To answer this question, we computed the means for each language/dataset pair and repeated the clustering process.
As expected, for most languages, mean vectors from all datasets that contain the language clustered together. Only for 5 languages of the 47 that appear in at least two training datasets, the datasets were separated in two clusters. For 4 of those cases (Tagalog, Levantine Arabic, Urdu/Hindi and Pushto), one of the clusters corresponded to the RATS data. As explained in Appendix \ref{app:train_data}, RATS data is telephone data retransmitted through different channels, some of which severely degrade the audio quality. Yet, RATS contains 38 languages, all of which appear in at least one other training dataset. The fact that, from these 38 languages, only 4 of them do not cluster well with the same language from other datasets suggests that it is the language and not the acoustic condition that makes the RATS data from these 4 languages different. We hypothesize that this data might have labelling issues or heavy code switching. 

The clusters in Figure \ref{fig:clusters}  correspond closely with expected groupings of languages based on both genetic/historical relationships as well as cultural and region-based ``sprachbund'' associations. For example, the cluster from Marathi to Sindi groups languages in Northern India, as one would expect based on their strong genetic relationship, but it places this cluster in close proximity to the cluster from Tamil to Kannada in Southern India. These two clusters share no known genetic relationship and are in separate language families (Indo-European and Dravidian) but there is a clear and close cultural and regional association, with influences on multiple levels spanning lexicon, phonemic systems and prosody. We see this again in the cluster spanning Galician to Basque, in which the Romance languages of Spain cluster tightly with one another, but also with Basque, which has a great deal of regional and historical contact with Iberian Romance languages impacting vocabulary, prosody and phonology but is itself genetically unrelated. 
It is interesting to see that Hebrew is grouped with Yiddish within a cluster of Germanic languages, as opposed to other Semitic languages. This is likely due to the very prominent historical population of Yiddish speakers and their descendants that make up a large part of the speakers of Modern Hebrew, with sharing of vocabulary, grammar and sound structure, similar to how Irish has a strong impact on the English of Ireland, for example.
Another cluster stands out as especially strange, showing the closest (though still relatively distant) neighbour of English as Maori, a language family of New Zealand. This is likely also due to cultural contact and large amount of borrowings from English to Maori, as well as actual code switching in the data that is natural when the vast majority of Maori speakers are also fluent in English. Other sprachbund effects are present in the clustering of Breton (Celtic) with French and Occitan (Romance). Most of the clusters are surprisingly precise and correspond closely with human intelligibility and intuitions (Polish/Czech/Slovak, Russian/Belorussian/Ukrainian, Dutch/Africaans, Icelandic-Danish group, etc.) and, with very few exceptions, group languages as one would anticipate from a linguistic perspective.

\subsection{System Configuration}
\label{sec:configuration}

The discriminative PLDA backends, both standard and hierarchical, were trained using Adam optimization with batches of size 2048, using the same 3-stage procedure and hyperparameters described in detail in \cite{ferrer2022speaker}. The first stage consists on training the model over 12000 batches with a learning rate of 0.0005. Then, the learning rate is increased to 0.001 and the model is trained with 3000 more batches. The best model from this stage, based on the average loss on the 7 held-out sets from training plus BABEL for the 4-, 8-, and 16-second chunks, is then further fine-tuned with a learning rate of 0.00001. The batches are created by randomly selecting the same number of samples for each language in each training dataset where the language is available. We found this method to be better than simply balancing out by language, since it allows the model to take better advantage of the multiple training datasets. The value of $\pi$ is set to 0.01. The optimal value of $\alpha$ on the datasets held-out from training was around 0.2. Yet, we found that this value was suboptimal for Babel data, where a value of 0.0 was better. Hence, we selected the latter value, assuming it would generalize better to other unseen conditions.
For each model, we ran the training procedure with 5 seeds and selected the best one based on the average performance on the development sets. 

The LDA dimensions were set to the maximum possible given the number of input classes, impossed by the maximum likelihood estimates we used for initialization. In the case of the standard non-hierarchical version, this corresponds to the number of languages minus 1, that is, 99 dimensions. In the case of the hierarchical version, the LDA dimension for the first PLDA stage is given by the number of language clusters minus 1, which is 71 in our case (note that languages that are not clustered with any other are still considered as a separate cluster). For the second stage, the maximum LDA dimension is given by the number of languages minus the number of clusters, since, at initialization, the embeddings from each cluster are centered by the mean of the per-language means, which reduces the rank of the between-class covariance matrix by one for each cluster. This corresponds, in our case, to 28 dimensions ($100-72$). Note that, given these LDA dimensions, the standard model has 67819 parameters, while the hierarchical model has 85078, a 25\% increase in the number of parameters. Most of the additional parameters in the hierarchical model are due to the $m_c$ values (72 vectors of size 384) that are subtracted from the embeddings for each cluster.

For the standard PLDA backend, samples were weighted when estimating the LDA and PLDA parameters, as explained in \cite{ferrer2022speaker}. The weights were defined to be proportional to the inverse of the number of samples for each language and dataset, simulating the behaviour obtained when generating the balanced batches in discriminative training. Note that the parameters for the standard PLDA backend coincide with the initial parameters for the non-hierarchical discriminative PLDA backend. Yet, for the standard PLDA  backend, we computed the scores using the exact LR formulation rather than the approximated one where each language is represented by a single enrollment vector ignoring the number of available samples. This gave consistent improvements for this backend. On the other hand, the exact scoring formulation did not appear to make a difference after discriminative training. We tested this using a variant of the formulation described in Section \ref{sec:hier}, where rather than discriminatively training the parameters in Equation \ref{eq:plda_score}, we train the parameters $B$, $W$ and $\mu$, which allows us to compute the score using the correct formulation. This gave very similar results, after discriminative training, compared to the simpler and  faster approach of training $\Gamma$, $\Lambda$, $c$ and $k$ (Equation \ref{eq:plda_score}) using the approximate formulation for scoring.

\subsection{Performance Metrics}

When testing, every sample is tested against every language detector in the model, creating $L$ scores per sample, where $L$ is the number of language detectors in the model. We measure performance in terms of actual detection cost function (DCF) \cite{van2007introduction} over those scores, with cost of miss and false alarms of 1 and probability of target of 0.1. That is, we report DCF~=~0.1~Pmiss~+~0.9~Pfa, where Pmiss is the percentage of positive trials labelled by the system as negative and Pfa is the percentage of negative trials labelled as positive. The threshold used to make decisions based on the LLRs is given by the theoretically optimal one given by Bayes decision theory for the chosen cost function.
We also compute the minimum value for the DCF where the threshold is selected to optimize the cost value. The difference between the actual and the minimum DCF indicates the amount of misscalibration present in the scores (see, for example, \cite{van2007introduction}).
We normalize the DCF value, as commonly done in NIST evaluation, dividing it by the value it would take for a non-informative system that always chose the class that led to the minimum cost. Hence, the DCF of a non-informative but well-calibrated system would be 1.0.

Note that we decided not to balance out the languages when computing the actual DCF, as done by NIST in the evaluations \cite{evalplanLRE17}, since many of the datasets are quite imbalanced in terms of languages, which generates very noisy cost functions when the infrequent languages are weighted equally to the most frequent languages. On the more balanced datasets (like LRE15 and LRE17), the balanced and imbalanced versions of the metrics are very similar. 

Confidence intervals are computed for the actual DCF using the bootstrapping method. Test waveforms are sampled with replacement 1000 times and, for each resulting set, the actual DCF is computed. Then, the 2.5 and 97.5 percentiles of the resulting empirical distribution for the DCF (composed of 1000 values) are used as the confidence interval for the DCF. Finally, while equal error rate (EER) is not our main metric of interest since it does not reflect calibration performance, it is a widely used metric for this task. Hence, we include EER values in an appendix for further reference.

\section{Results and Discussion}

\begin{figure*}
\centering
\includegraphics[height=5.3cm]{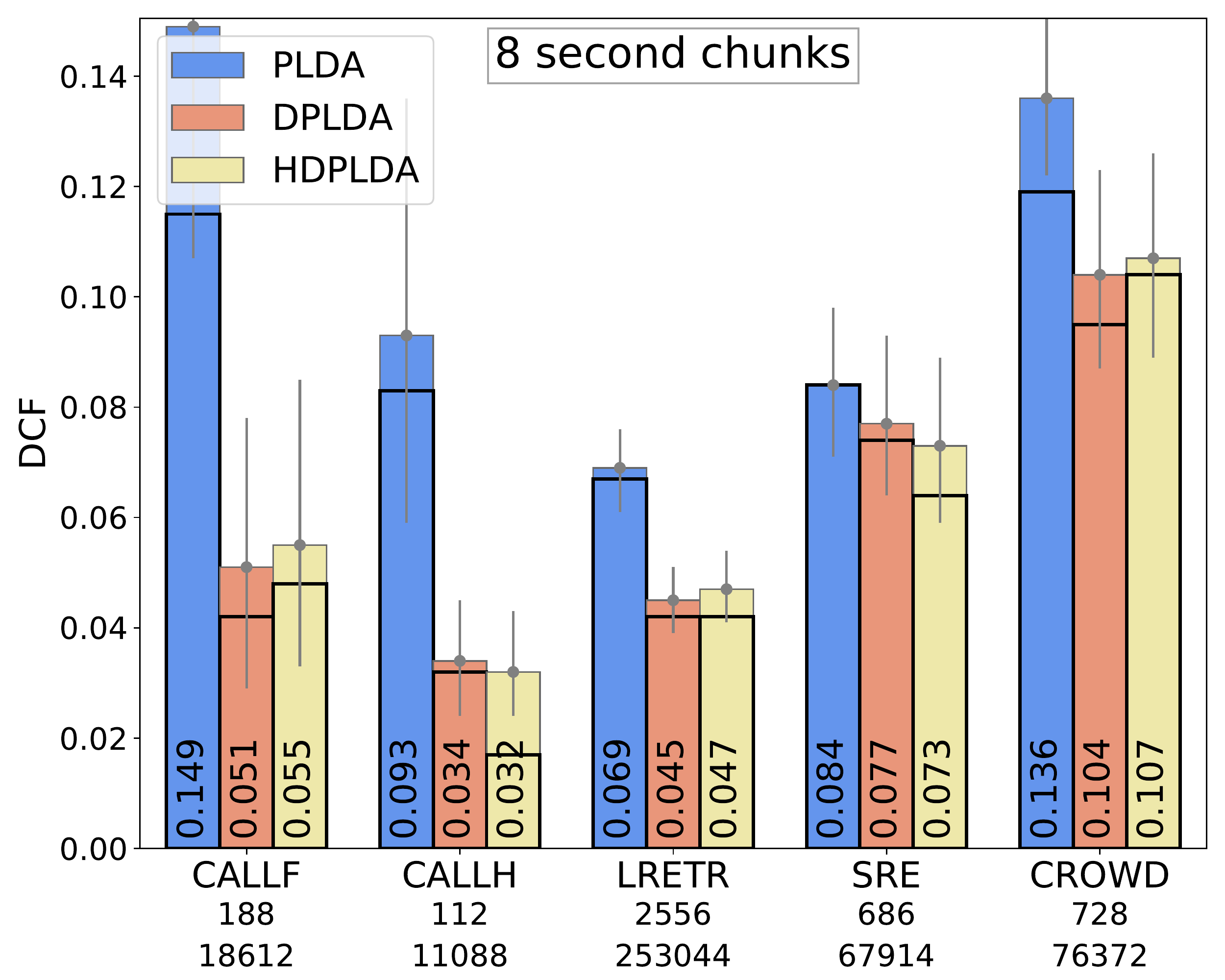}
\includegraphics[height=5.3cm]{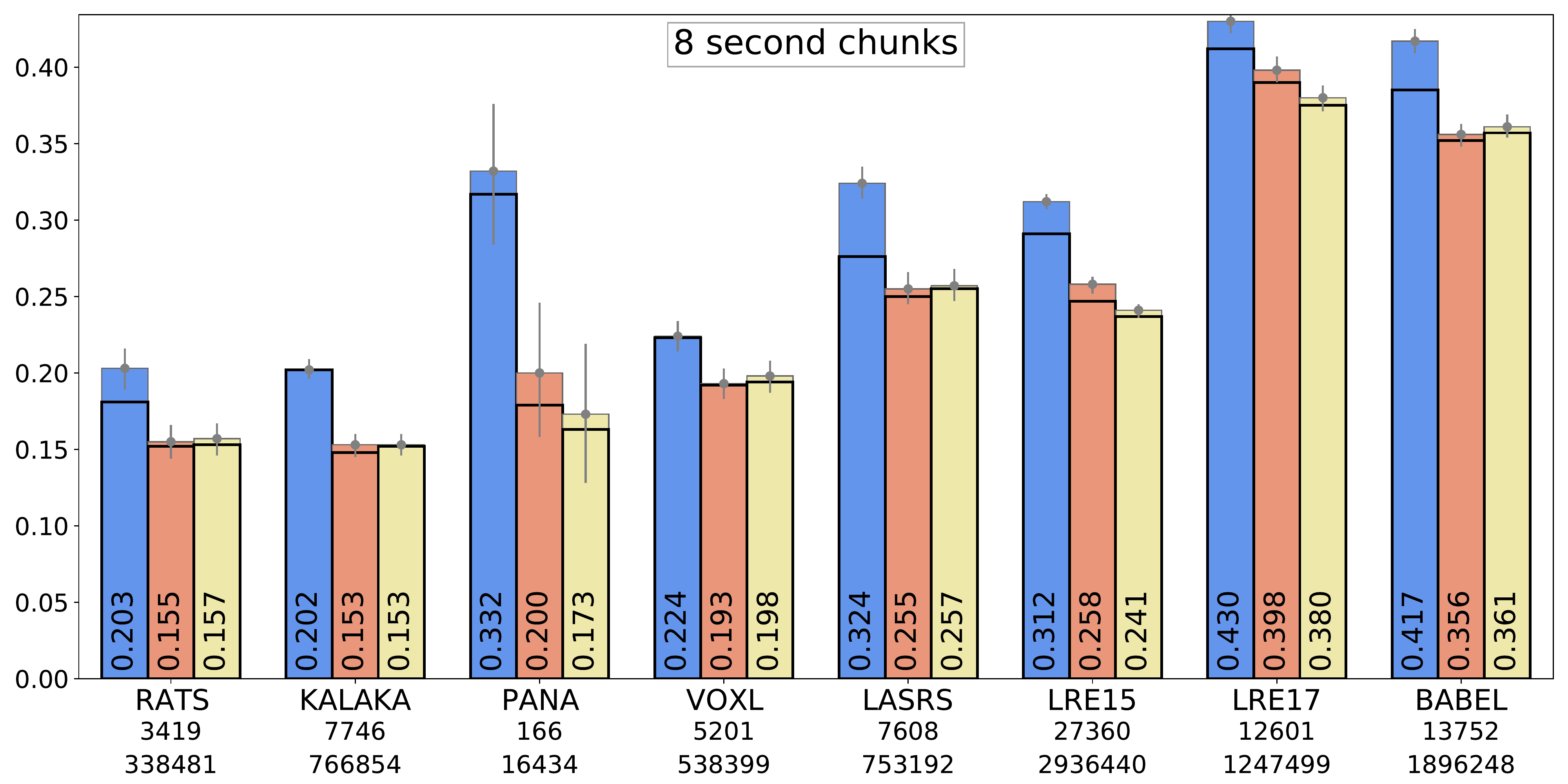}
\includegraphics[height=5.3cm]{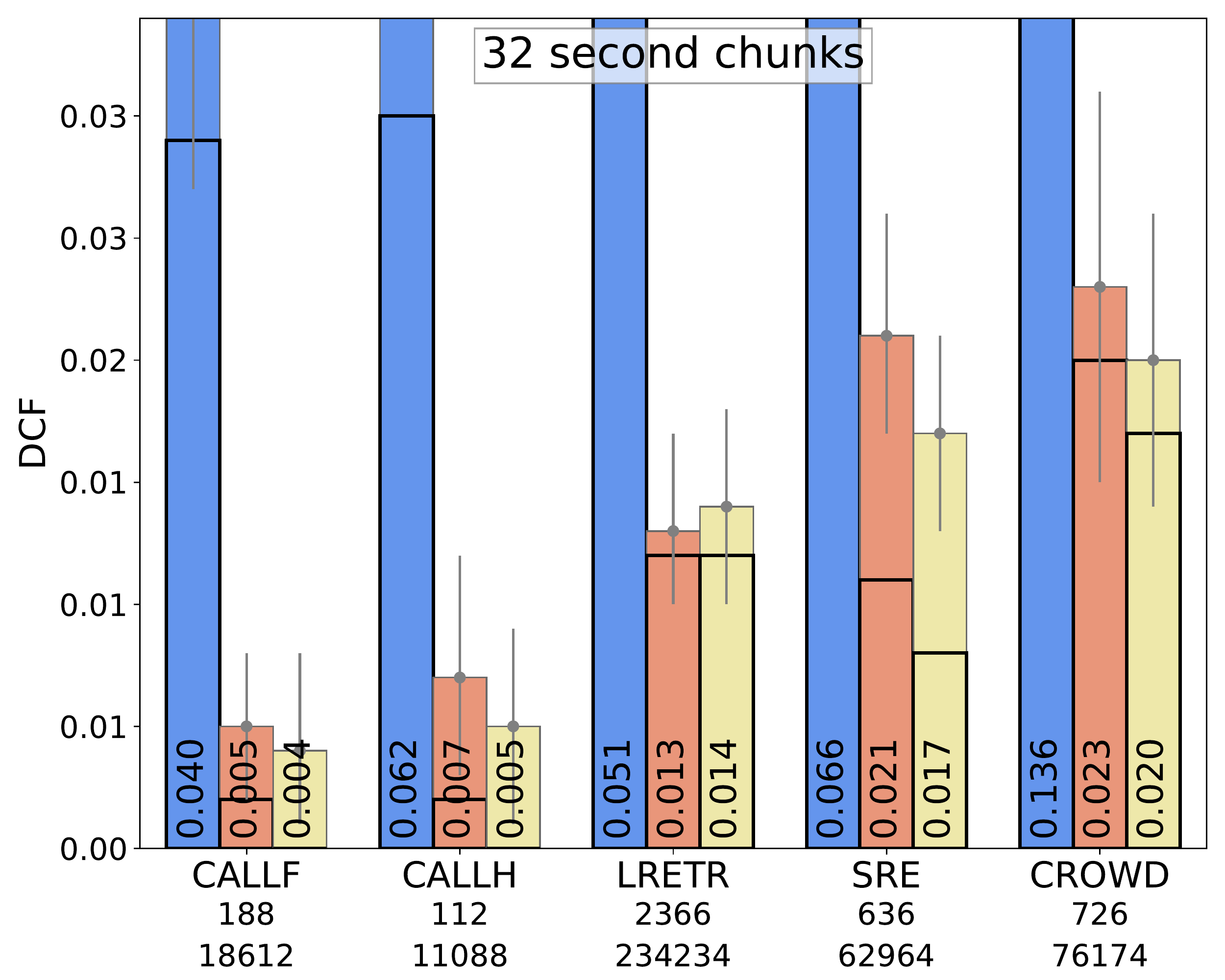}
\includegraphics[height=5.3cm]{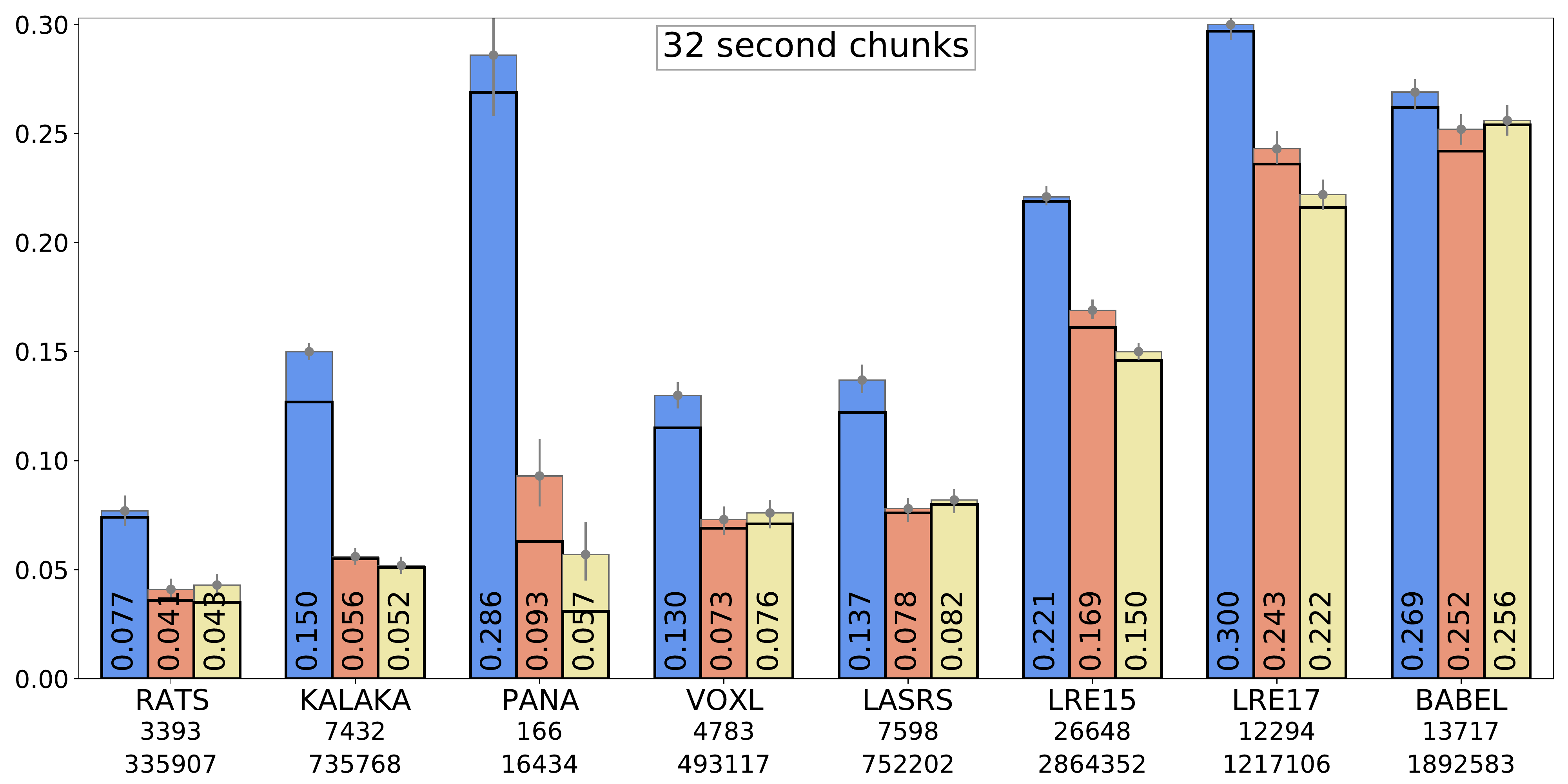}
\caption{DCF results for the development (CALLH, CALLF, LRETR, SRE, PANA, RATS, VOXL, BABEL) and evaluation sets (CROWD, KALAKA, LASRS, LRE15, LRE17) chunked to 8 and 32 seconds, for PLDA, DPLDA and HDPLDA. The numbers inside the bars correspond to the height of the bar, the actual DCF. The black line inside each bar indicates the value of the minimum DCF. The grey vertical lines indicate the confidence intervals for the actual DCF. Results are split in two for each duration, to adjust the y range to each subgroup. Sets are ordered in terms of increasing DCF for HDPLDA on the 32-second chunks. The values under the set names are the number of positive (top) and negative (bottom) trials in each set.}
\label{fig:eval}
\end{figure*}

Figure \ref{fig:eval} shows the actual and minimum DCF results for the following three backends: 
\begin{itemize}
    \item {\bf PLDA}: the standard PLDA backend with parameters trained by maximizing likelihood,
    \item {\bf DPLDA}: the discriminatively-trained version of the PLDA backend, and
    \item {\bf HDPLDA}: the hierarchical discriminatively-trained backend proposed in this work.
\end{itemize}
As explained above, test samples are chunked to different durations for testing. 
Results are shown for two chunk durations of 8 and 32 seconds. Average DCF over all datasets in Figure~\ref{fig:eval} for the 32-second chunks can be found in Table \ref{table:avres} (All-trial column). EER values for all datasets and systems are included in Appendix \ref{sec:eer_results}. 

Figure~\ref{fig:eval} shows that, with few exceptions, discriminative training consistently improves performance of the PLDA backend, specially for longer durations. The hierarchical approach shows modest gains on some of the test datasets over the flat DPLDA approach. 
As we will see next, the gains cannot be clearly appreciated on these results because the main advantage of the HDPLDA method over the DPLDA method is seen when focusing on closely related languages. 
As a side note, the average DCF over all evaluation datasets for HDPLDA is approximately 7\% worse when the backend is trained without augmented samples (without changing the embedding extractor which is the one described in Section \ref{sec:embeddings} and is trained with augmentations).

 \begin{table}[!t]
\caption{Average DCF  over all test sets in Figure \ref{fig:eval} (second column, all trial results) and Figure \ref{fig:eval_clust10} (third column, by-cluster results) for the 32-second chunks, for the same three systems in the figures, plus a smaller version of the HDPLDA system.\label{table:avres}}
\centering
\begin{tabular}{l|r|r}
System & All-trial & By-cluster \\
\hline
PLDA & 0.148 & 5.89 \\
DPLDA & 0.082 & 1.03 \\
HDPLDA & 0.077 & 0.79 \\
HDPLDA small & 0.084 & 0.77 \\
\end{tabular}
\end{table}

\begin{figure}
\centering
\includegraphics[width=\columnwidth]{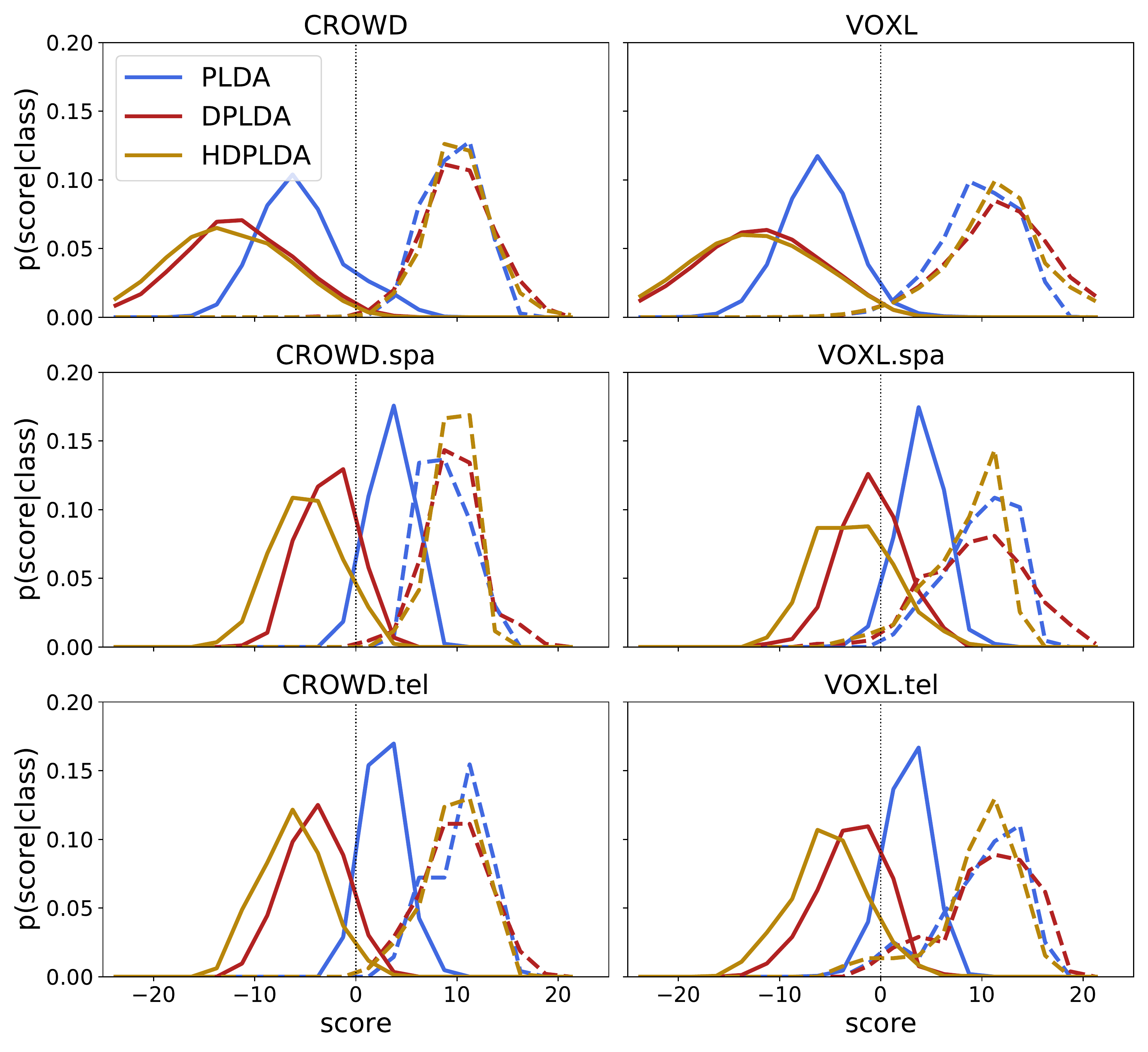}
\caption{Score distributions for the same three systems as in Figure \ref{fig:eval} for the CROWD and VOXL sets on all trials (top row), and for two trial subsets: the spa cluster trials (middle row) and the tel cluster trials (bottom row). Solid and dashed lines correspond to the negative and positive class score distribution, respectively.}
\label{fig:score_dist}
\end{figure}

To explore the behaviour of the different approaches on closely related languages, Figure~\ref{fig:score_dist} shows the distribution of scores per class for the same three systems shown in Figure \ref{fig:eval} for two different datasets, CROWD and VOXL. The first row shows the distribution over all trials (i.e., all samples tested against all available detectors); the same ones used to compute the results in Figure~\ref{fig:eval} and the All-trial column in Table~\ref{table:avres}. The distributions show, qualitatively, the same results as the DCF: both DPLDA and HDPLDA outperform PLDA (i.e., negative and positive distributions overlap less), while performing very similar to each other. The second and third row show the scores on a subset of positive and negative samples, corresponding to two clusters, spa and tel. The languages included in these clusters can be found in Table \ref{table:clusters}. In these distributions, both the test samples and the detectors are restricted to the languages in the corresponding cluster. We can see that the positive distribution for the within-cluster trials is very similar to the overall distribution (top row in the figure). This is expected, since those trials should not be, a priori, any harder or easier than those for any other language. On the other hand, the negative within-cluster trials are particularly hard trials, since they correspond to samples of one language in the cluster evaluated against a detector from another language in the same cluster. As a consequence, the negative score distribution for the within-cluster trials drastically moves to the right with respect to the distribution on all trials. While this happens  for all three systems, these plots again show very clearly the advantage of DPLDA and HDPLDA over PLDA: their negative distributions stay further to the left and overlap the positive distribution much less than for PLDA. Further, comparing DPLDA with HDPLDA we can see a consistent advantage from HDPLDA in that the negative distribution is further to the left than for DPLDA. 

 \begin{table}[!t]
\caption{Language clusters with at least two languages. The name for each cluster is chosen as the ISO code for a randomly chosen language from the cluster. \label{table:clusters}}
\centering
\begin{tabular}{l|l}
Name & Languages \\
\hline
arz & Egyptian, Gulf, Levantine, Mesopotamian Arabic \\
slk & Slovak, Czech, Polish \\
spa & Spanish, Catalan, Galician \\
tat & Tatar, Bashkir, Kazakh \\
tel & Telegu, Tamil, Malayalam, Kannada \\
ukr & Ukrainian, Belarusian, Russian \\
\end{tabular}
\end{table}

\begin{figure*}
\centering
\includegraphics[height=5cm]{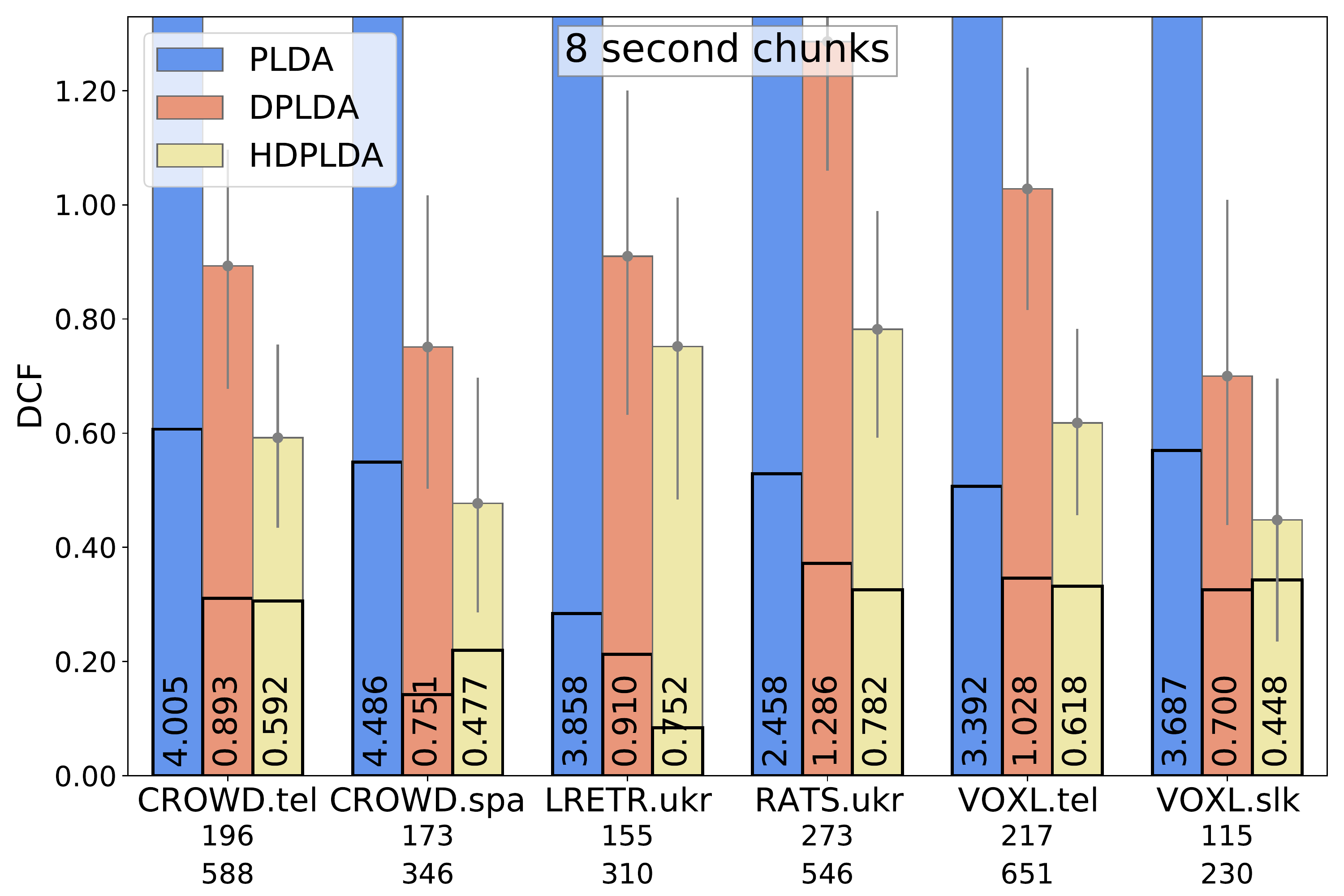}
\includegraphics[height=5cm]{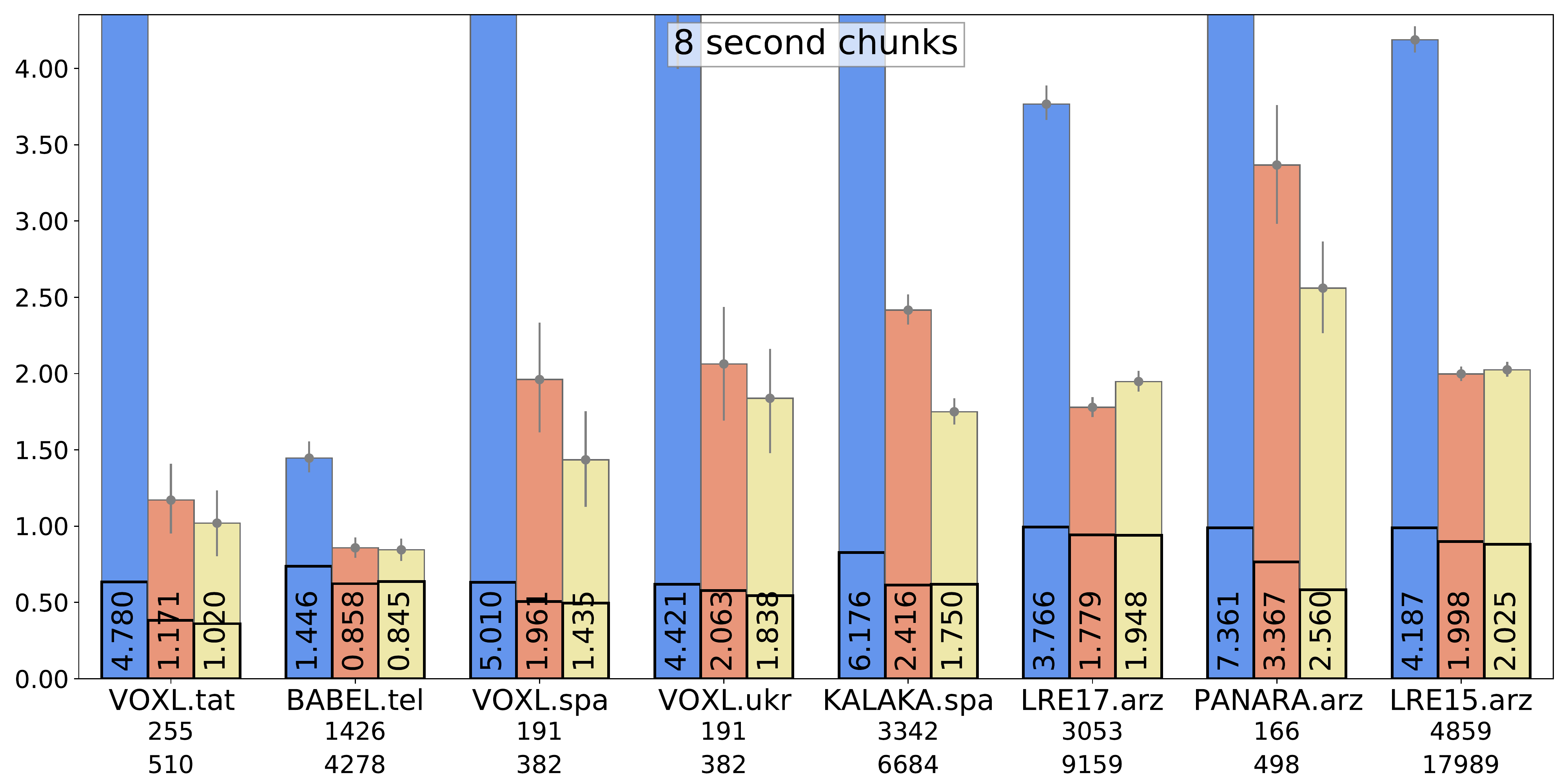}
\includegraphics[height=5cm]{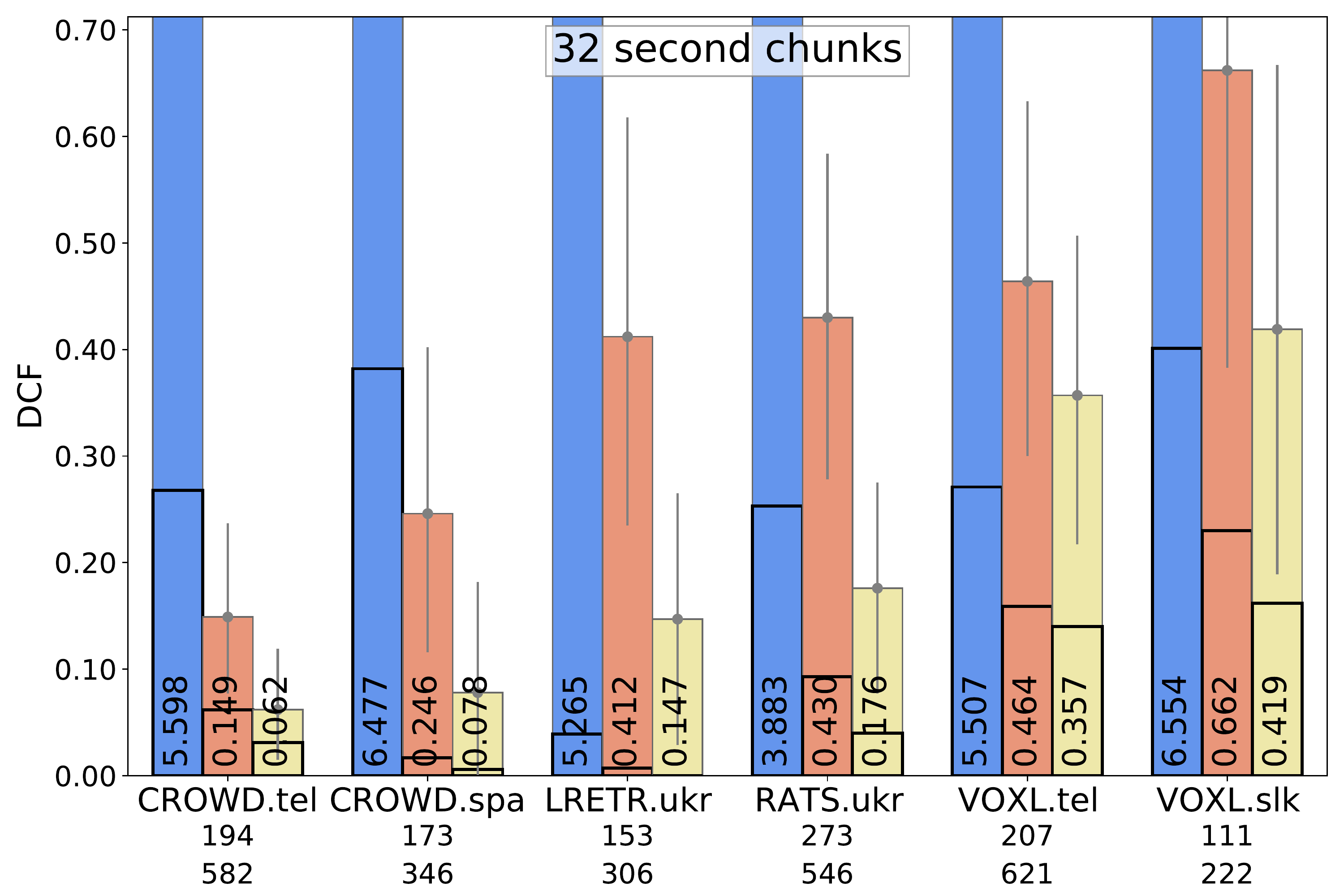}
\includegraphics[height=5cm]{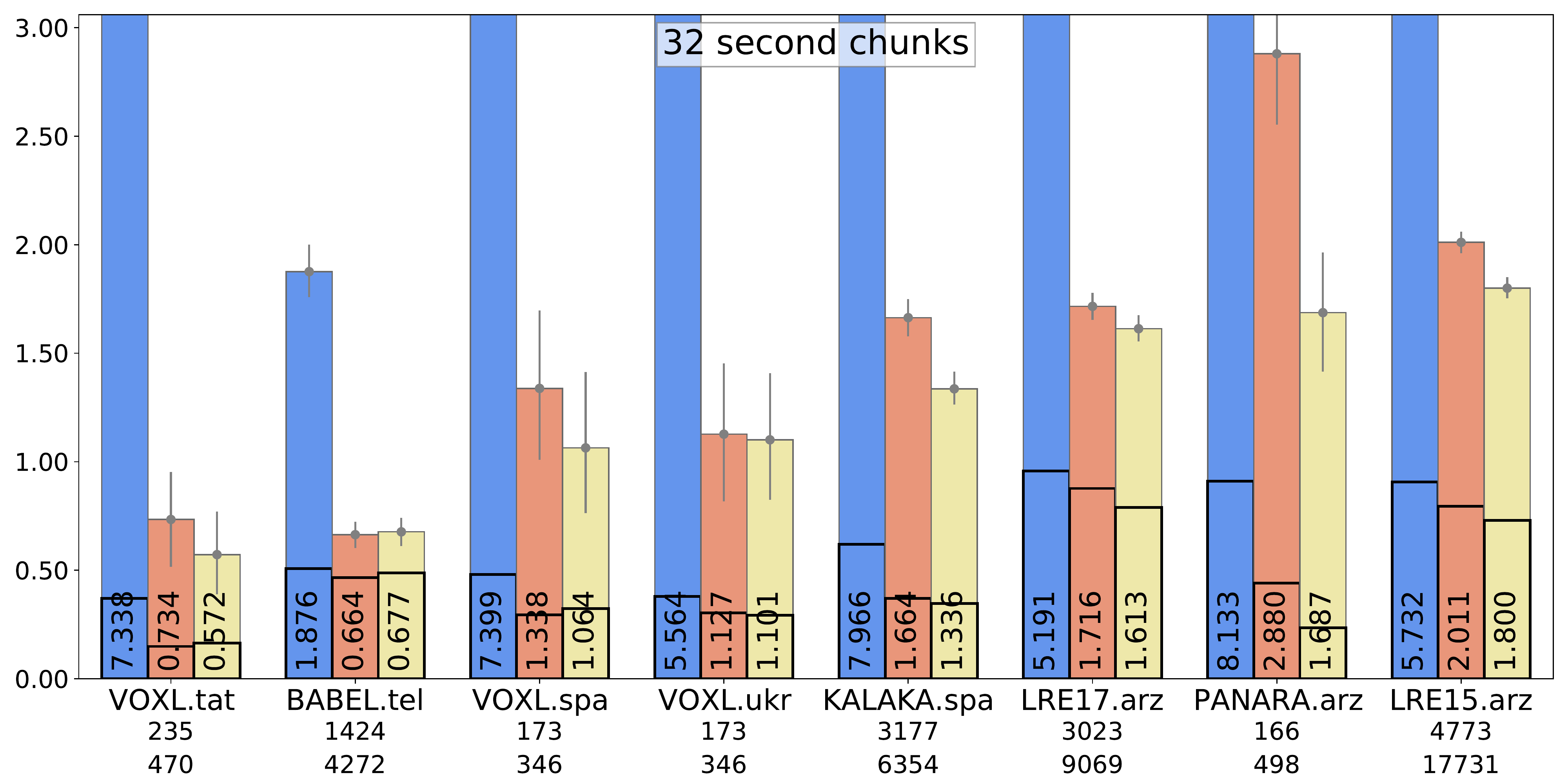}
\caption{Same as Figure \ref{fig:eval} but with results computed over specific clusters using a threshold of 10. The cluster is identified by the name of one of its language (see Figure \ref{fig:clusters} to find the list of languages in each cluster).}
\label{fig:eval_clust10}
\end{figure*}

The distributions in Figure \ref{fig:score_dist} suggest that HDPLDA has better performance than DPLDA on the within-cluster trials. 
A more quantitative way to make this analysis is to compute the DCF on the within-cluster trial subsets. Figure \ref{fig:eval_clust10} shows the results when using the same clusters used to train the model, selecting only the clusters with at least three languages (Table~\ref{table:clusters}) and, for each cluster, only the datasets which have samples for at least two languages in that cluster.  Average DCF over all datasets in Figure~\ref{fig:eval_clust10} for the 32-second chunks can be found in Table \ref{table:avres} (By-cluster column). EER values for all datasets and systems in this Figure are included in Appendix \ref{sec:eer_results} for further reference.

We can see that the qualitative observation made from Figure \ref{fig:score_dist} -- DPLDA and HDPLA are markedly better than standard PLDA, and HDPLDA is better than DPLDA -- is confirmed by the DCF values in Figure~\ref{fig:eval_clust10} and Table~\ref{table:avres}. Further, conclusions are consistent across datasets and clusters, with only a few exceptions where HDPLDA does not significantly outperform DPLDA. Interestingly, the gains are generally larger for the longer chunks, suggesting that, when more information is available in the signal, HDPLDA can take better advantage of it than DPLDA. 

While the gains from HDPLDA in the within-cluster results are large, these gains show up as much smaller relative gains when evaluating on all trials (Figure \ref{fig:eval}). This is because only a very small percentage of all possible negative trials when using all detectors and all samples are difficult trials since most detectors correspond to languages that are not closely related to the languages in the samples. For example, for the case of CROWD, the total number of negative trials for the 32-second chunks in Figure \ref{fig:eval} is 76174, of which only 582 and 346 (less than 0.8\%) correspond to the hard trials for the tel and spa clusters, respectively. 

We note that we obtained similar conclusions to those in Figure \ref{fig:eval_clust10} when using language clusters that do not exactly coincide with those used in training. Yet, larger clusters include languages that are less related to each other and, as a consequence, the negative distributions start to look more like those for all trials resulting in smaller relative gains from HDPLDA. Clearly, the biggest benefit from HDPLDA lies in its superior ability to differentiate very closely related languages compared to the non-hierarchical approaches.

A notable observation from Figure \ref{fig:eval_clust10} is that while HDPLDA greatly improves the DCF over plain PLDA for the within-cluster trials,  even for this system, there is a large difference between the minimum and the actual DCF. This is reasonable since the system was trained on all available languages at once, resulting on well-calibrated scores when all detectors are used (as in Figure \ref{fig:eval}). On the other hand, when only a subset of closely-related language detectors are used, scores become poorly calibrated. Note, though, that the goal of this analysis is not to show results for a scenario with restricted language detectors in which case the LLRs would have to be transformed to adapt to the new set of detectors, but to highlight the gains from HDPLDA which are hidden in the overall results.

As explained in Section \ref{sec:configuration}, the HDPLDA backend contains 25\% more parameters than the DPLDA and PLDA backends. We may then wonder whether the gains are due specifically to the proposed architecture or simply to the fact that the model now has more parameters. In our approach, the size of the DPLDA backends cannot be increased to make it comparable to HDPLDA because it is limited by the number of LDA dimensions that can be estimated with the available number of training languages. On the other hand, we can decrease the size of the HDPLDA model to make it comparable to that of the DPLDA model. Reducing the size of the by-cluster LDA transform from 71 to 50 results in a model with 70378 parameters, comparable to the size of the DPLDA model. The last line in Table \ref{table:avres} shows the results for this reduced HDPLDA system. We can see that, while the smaller HDPLDA system
degrades slightly with respect to the large one when evaluating on all trials, it still shows a large gain with respect to DPLDA on the by-cluster results. 
This suggests that it is not the increase in number of parameters that is giving an advantage in the model's ability to differentiate closely-related languages but that the proposed architecture plays an essential role.

The task considered in this paper is the creation of a multi-purpose language detector that can be used out-of-the-box on many different scenarios and conditions to detect a large number of languages. We have shown that, for this task, discriminative training and, to a lesser extent, the proposed hierarchical approach, lead to gains over the standard PLDA approach. In the future we plan to explore whether this general-purpose system can also be used for more specific tasks where only a subset of languages is of interest. As shown here, the scores generated by the system for those cases are misscalibrated since they were trained for a different task. In order to compute well-calibrated likelihood ratios for any subset of detectors, the model needs to output scaled likelihoods. For this reason, in the future we plan to explore discriminative and hierarchical models that output likelihoods instead of LLRs.
Finally, while the experiments in this paper are done for one specific type of input feature and embedding extractor, we believe the conclusions would generalize to other types of features and embeddings. We base this hypothesis partly on related results that show that DPLDA gives similar gains over PLDA for the speaker verification task when using two different types of embeddings \cite{Estevez:2022}. Nevertheless, we plan to explore different types of embeddings in our future work.

\section{Conclusions}

We presented a novel hierarchical approach for language detection, which we call HDPLDA, where the computation of the log-likelihood ratios (LLRs) for each language is made in two separate stages: one that computes LLRs for language clusters and another one that computes LLRs for the languages conditional to their cluster. Each of the two stages has the same functional form as a standard PLDA backend. The model is trained discriminatively to optimize the average binary cross-entropy over all language detectors. We compared the proposed approach with a standard PLDA backend and with a discriminatively-trained version of such backend (DPLDA). The systems were trained on several datasets which include over 100 languages, and tested on a large variety of datasets, including conditions matched and mismatched to those in training.  We showed that the DPLDA system outperformed PLDA on all datasets by a large margin, specially on longer speech segments and easier conditions. Further, we show that HDPLDA outperforms DPLDA by improving the model's ability to reject languages that are highly related to but different from the language in the test sample.

While our training dataset contains a large number of languages, few of them are closely related to each other. In particular, only 6 language clusters contain at least 3 languages. We believe that the proposed hierarchical approach would result in a larger relative improvement if trained on a larger number of languages including more and larger language clusters. We plan to continue expanding our training dataset to explore this hypothesis.
Finally, we hope that the code, example scripts and configuration files provided in \url{https://github.com/luferrer/DCA-PLDA} will allow other groups to test the method on their own embeddings and datasets.

\section{Acknowledgments}

We thank Sarak Bakst, Alex Erdmann and Chase Adams for their help on linguistic issues related to data labelling.

\appendices
\section{Training Data}
\label{app:train_data}

\begin{table}[!t]
\caption{Number of languages, and number of training samples for each  dataset used in training. \label{table:training_stats}}
\centering
\begin{tabular}{l|r|r}
SET & \#Lang & \#Samples   \\
\hline
CALLF & 12 & 971  \\
CALLH & 5 & 556 \\
LRETR & 47 & 69957 \\
PANA & 4 & 862 \\
RATS & 38 & 98187 \\
SRE & 16 & 43343 \\
VOXL & 89 & 34584 \\
\hline
TOTAL & 100 & 248460\\
\end{tabular}
\end{table}

Our training data is composed of 7 datasets:
\begin{itemize}
\item {\bf CallFriend (CALLF) and CALLHOME (CALLH)} \cite{callhome} were developed by the Linguistic Data Consortium (LDC), consist of telephone calls in several different languages, and are available for download from the LDC as separate corpora for each language or dialect.
\item {\bf LRETR} is composed of several datasets that were used at SRI for training and evaluation of the systems developed for the 2009 language recognition evaluation (LRE) organized by NIST. It includes LRE data for the years 2003, 2005, 2007 and 2009 \cite{martin2003nist,martinLRE05,Martin:2009} and the Voices of America dataset, released as training data for the 2009 LRE \cite{evalplanLRE09}.
\item {\bf Panarabic (PANA)} (citation not available) was collected by Appen on-site in the Middle East and is focused on five regional variants of Arabic: Egypt, Gulf, Palestine, Syria and Iraq. The goal of the data was to have highly accurate and verified spontaneous examples of each dialect without potential channel dependencies and other flaws for purposes of dialect detection research. Each dialect was represented by 100 speakers, each of whom spoke for approximately 30 minutes. This data was collected in-person from two-party, face-to-face conversations on a random set of topics in a relatively controlled environment. Collection apparatus consisted of two channels, close talking label mic and distant table mic and conversants where monitored to ensure they spoke the expected regional dialect during the collection process. Data was saved as 8-bit A-law with 8 Khz sample rate.
\item {\bf RATS}~\cite{ref:rats_set} is composed of approximately 350 hours of telephone conversations. The audio was retransmitted over eight channels, making 3,000 hours of total audio. 
\item {\bf SRE} consist of data from the NIST Speaker Recognition Evaluations from 2004 to 2019 \cite{nist_sre0406,nist_sre08,nist_sre10,nist_sre12,nist_sre16,nist_sre18,nist_sre19,greenberg2020}, which includes telephone and microphone recordings for telephone conversations and interviews. 
\item {\bf Voxlingua107 (VOXL)} \cite{valk2021slt} consists of short speech segments automatically extracted from YouTube videos and labelled according the language of the video title and description, with some post-processing steps to filter out false positives. For our work, we concatenate all the segments that correspond to the same video into a new longer waveform which is then later chunked into the durations we need for training and testing. This dataset is known to contain 2\% labelling errors based on crowdsource verification made by the authors. 
\end{itemize}

Two of the training sets, RATS and VOXL contain some extremely degraded samples, either due to channel distortion or to background noise or music. To prevent those samples from corrupting the model, we ran the YAMNet\footnote{\url{http://github.com/tensorflow/models/tree/master/research/audioset/yamnet}} model on all our training data and discarded any files for which the average probability of speech was lower than 0.3 or the average probability of music, chant, or sing labels was higher than 0.1. This procedure discarded around 30\% of the samples in RATS and VOXL and less than 5\% of the samples in all other datasets. Finally, we excluded from the training data, any languages for which less than 100 samples were available  after the filtering process described above (samples from those languages are still used in the held-out sets for testing as out-of-set samples). We also discarded Welsh data from VOXL, since in our preliminary error analysis we realized this data contained many samples that were spoken in English rather than in Welsh. Some of the languages discarded with this process were responsible for a large number of missed detections on the held-out sets in our earlier models trained with all the available data, indicating that, indeed, those detectors were not properly trained due to the large percentage of corrupted or mislabeled files. 
For some datasets, several chunks from a single recording are available. In these cases, we selected the longest chunk and discarded the rest, since we applied our own chunking process with a very specific distribution of durations. Table \ref{table:training_stats} gives the number of languages and samples used for training (i.e., after all filtering is applied) for each dataset. The break down of number of samples of each language for each dataset can be found in \url{https://github.com/luferrer/DCA-PLDA/blob/master/papers/hplda/}.

\section{Evaluation Data}
\label{app:eval_data}

The evaluation data is composed of 13 different sets. Table \ref{table:stats} shows the total number of languages, the number of samples and the proportion of out-of-set samples (i.e., samples that belong to languages not available in training) for each test set. Seven of these sets are composed of samples randomly held-out from each of the training sets. Note that for VOXL, the number of languages is larger in testing than in training, since some languages were discarded from training due to having too few samples but are kept in testing as out-of-set languages.
The other six test sets correspond to conditions not seen by the model during training and are described below.
\begin{itemize}
\item {\bf BABEL}~\cite{Gales:2014} is a multilingual dataset that consist of approximately 80 hours of speech of diverse languages from language packs released for the IARPA Babel program.  The data is recorded in real-life scenarios, such as conversational telephone speech, over a range of acoustic conditions, such as mobile phone conversation made from a car. This dataset is not used for training but it is used for hyperparameter, seed and epoch selection as a mismatched condition to those seen in training.
\item {\bf CROWD}~\cite{van2017rapid,sodimana2018step,gutkin-et-al-yoruba2020,he-etal-2020-open,demirsahin-etal-2020-open,kjartansson-etal-sltu2018,guevara-rukoz-etal-2020-crowdsourcing,oo-etal-2020-burmese,butryna2020google} consist of over 1,500 hours of speech of languages and dialects of South and Southeast Asia, Africa, Europe and South America. Data is freely available online hosted by Open Speech and Language Resources (\url{http://openslr.org}).
\item {\bf KALAKA} \cite{kalaka} is a dataset composed of TV broadcast speech for training, and audio data extracted from YouTube videos for tuning and testing. In this work, we use all this data for testing.
\item {\bf LASRS} is composed of 100 bilingual speakers from each of three languages, Arabic, Korean and Spanish \cite{Beck2004}. Each speaker is recorded in two separate sessions speaking English and their native language using several recording devices. Most speakers have a heavy non-native accent when speaking English.
\item {\bf LRE15 and LRE17} \cite{zhao2016results,sadjadi20182017} are composed of the data used for development and evaluation in the language recognition evaluations of 2015 and 2017. They include conversational telephone speech (CTS) data and broadcast narrowband speech (BNBS). 
\end{itemize}

 \begin{table}[!t]
\caption{Number of languages, number of samples and percentage of out-of-set (OOS) samples  for the 32-second chunks in each test set. The number of samples and percentage of OOS samples for 8-second chunks is very similar to the 32-second chunks in all cases so we do not include them here. The first block of sets are held-out from training datasets. The bottom block are conditions unseen during training. The 7 sets held-out from training plus BABEL are used for development. \label{table:stats}}
\centering
\begin{tabular}{l|r|r|r}
SET & \#Lang & \#Samples &  \%OOS  \\
\hline
CALLF & 12 & 188 & 0.0 \\
CALLH & 5 & 112 & 0.0 \\
LRETR & 47 & 2366 & 0.0 \\
PANA & 4 & 166 & 0.0 \\
RATS & 38 & 3393 & 0.0 \\
SRE & 16 & 636 & 0.0 \\
VOXL & 98 & 4979 & 3.9 \\
\hline
BABEL & 20 & 19063 & 28.0 \\
CROWD & 23 & 769 & 5.6 \\
KALAKA & 10 & 7432 & 0.0 \\
LASRS & 3 & 7598 & 0.0 \\
LRE15 & 16 & 28910 & 7.8 \\
LRE17 & 11 & 12294 & 0.0 \\
\end{tabular}
\end{table}

The break down of number of samples of each language for each dataset can be found in \url{https://github.com/luferrer/DCA-PLDA/blob/master/papers/hplda/}.

\section{Data Labelling Process}
\label{app:labelling}
Since datasets were labelled at different levels of granularity, we had to adopt a criteria to make the labels consistent across all datasets. We decided that languages that are not mutually intelligible should have different labels. Of course, this is a rather vague statement, since mutual intelligibility is not categorical. For this work, we defined two languages to be mutually intelligible if two speakers from those languages would be able to communicate somewhat fluently, though perhaps missing some words and with some extra cognitive effort due to differences in pronunciation. Under this criteria, we had to merge labels from several datasets. For example, LRE15 had three different Spanish labels for Latin America, European and Caribbean Spanish. According to our criteria, we merged those three sets into a single Spanish category. In other cases, a dataset had a label that was too wide for our criteria. For example, SRE Arabic data was annotated with a single label “Arabic". After listening to some samples, we found it included multiple regional variants of Arabic that are substantively different and are often treated as separate entities in the community, like Egyptian and Levantine Arabic. Since relabelling those samples with their finer-grained language/dialect labels would require a very large effort, we decided to discard all those samples from training and testing datasets.

We would like to note that the process of curating the different datasets to assign consistent labels and correct mistakes, as well as the process of discarding very corrupted samples (described in Appendix \ref{app:train_data}), was essential to obtain the performance shown in this paper. In particular, many of the errors we observed during early experiments were simply due to labelling issues or corrupted training data that resulted in corrupted models for those languages.

\section{EER Results}
\label{sec:eer_results}

\begin{figure*}
\centering
\includegraphics[height=5.3cm]{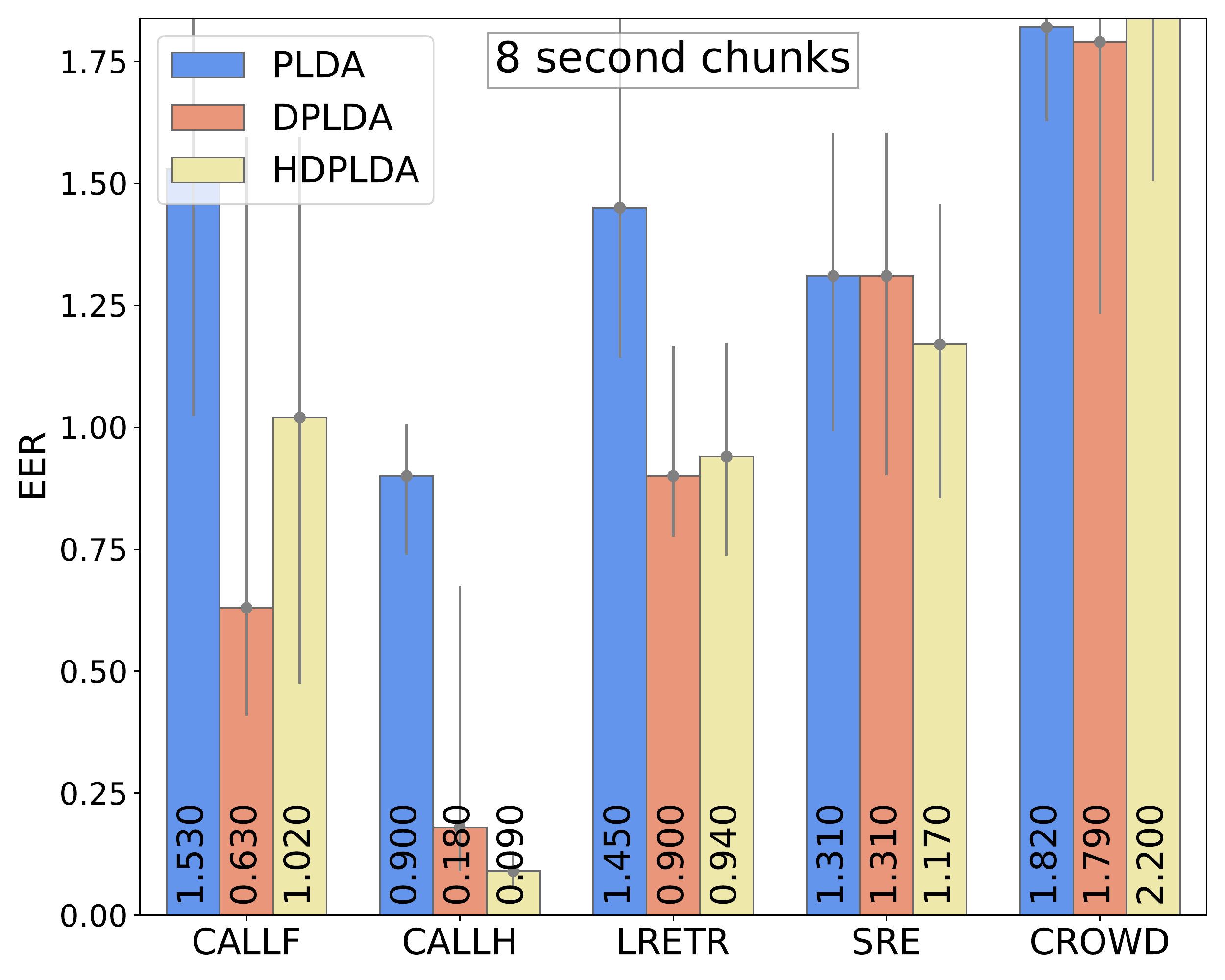}
\includegraphics[height=5.3cm]{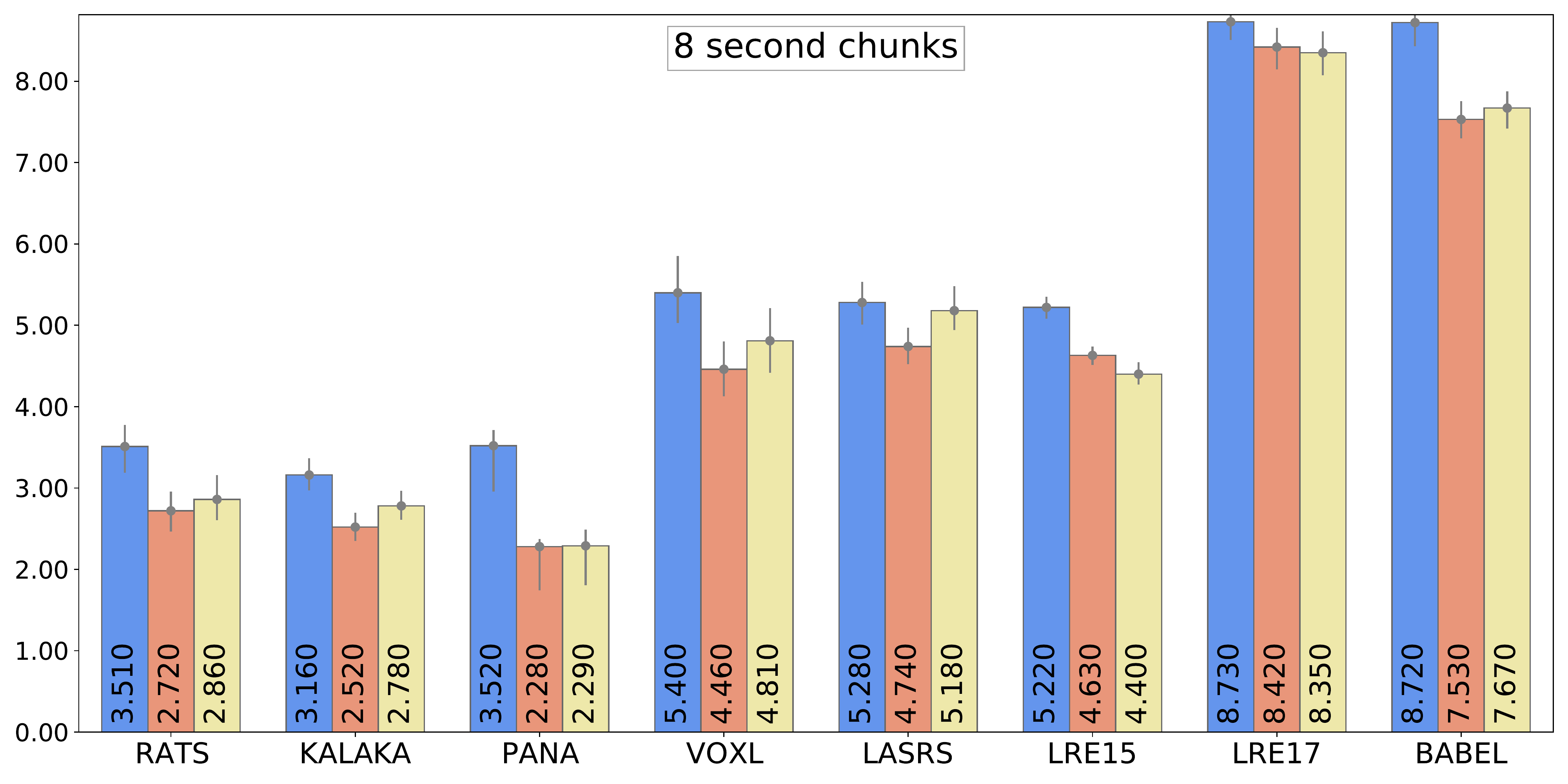}
\includegraphics[height=5.3cm]{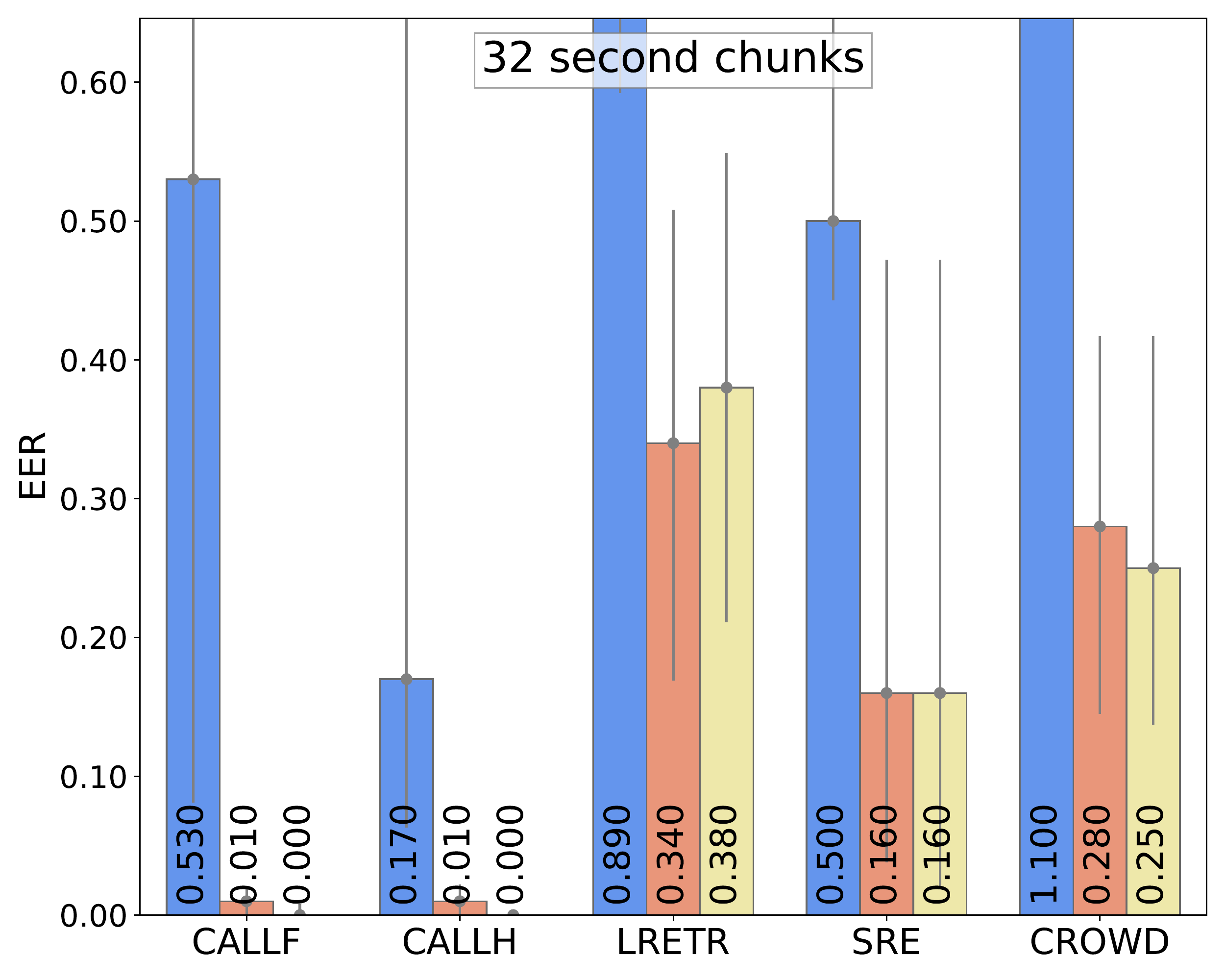}
\includegraphics[height=5.3cm]{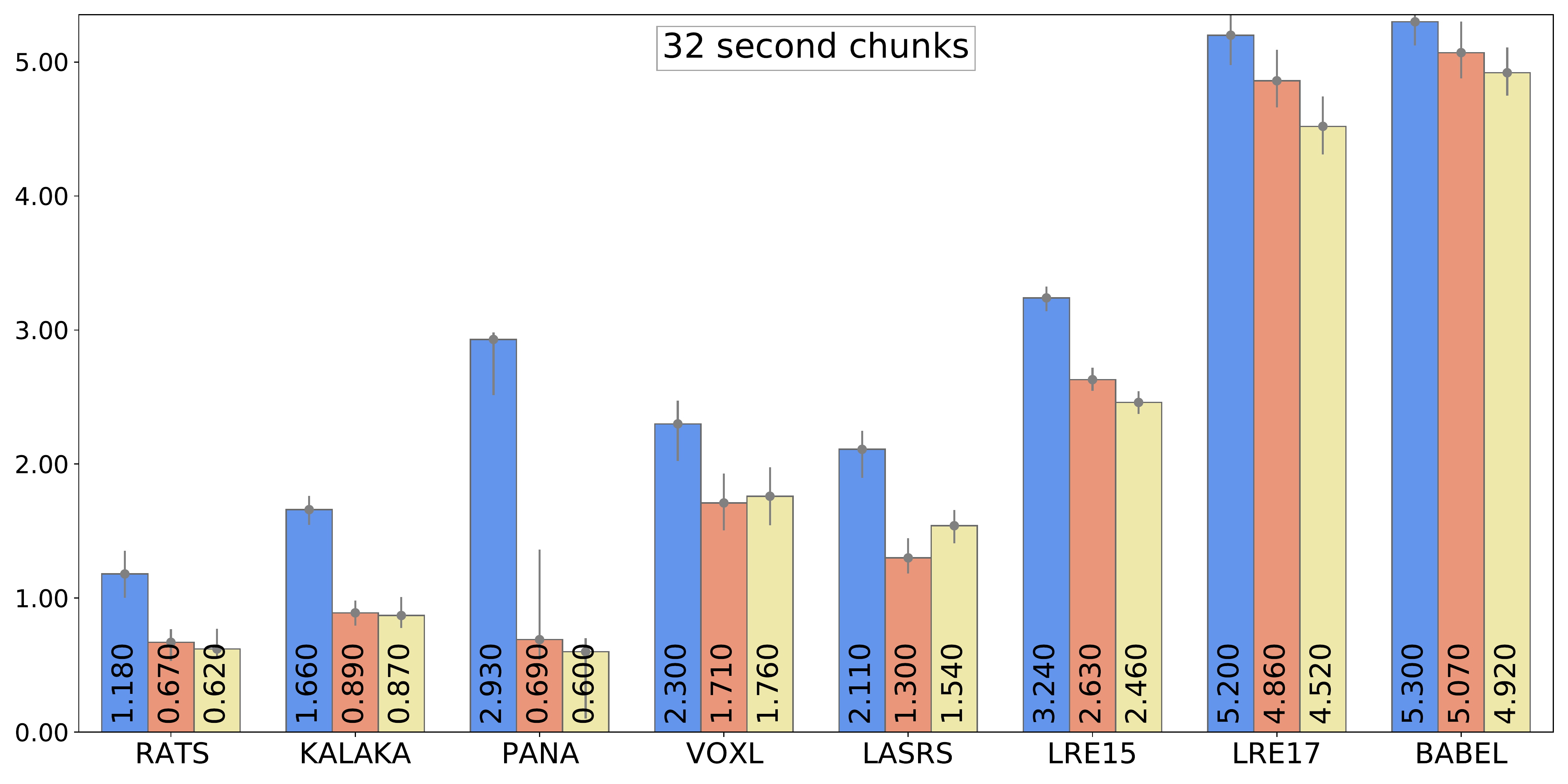}
\caption{EER results for the development (CALLH, CALLF, LRETR, SRE, PANA, RATS, VOXL, BABEL) and evaluation sets (CROWD, KALAKA, LASRS, LRE15, LRE17) chunked to 8 and 32 seconds, for PLDA, DPLDA and HDPLDA. The numbers inside the bars correspond to the height of the bar. Sets are split and ordered as in Figure \ref{fig:eval}.}
\label{fig:eval_eer}
\end{figure*}

\begin{figure*}
\centering
\includegraphics[height=5cm]{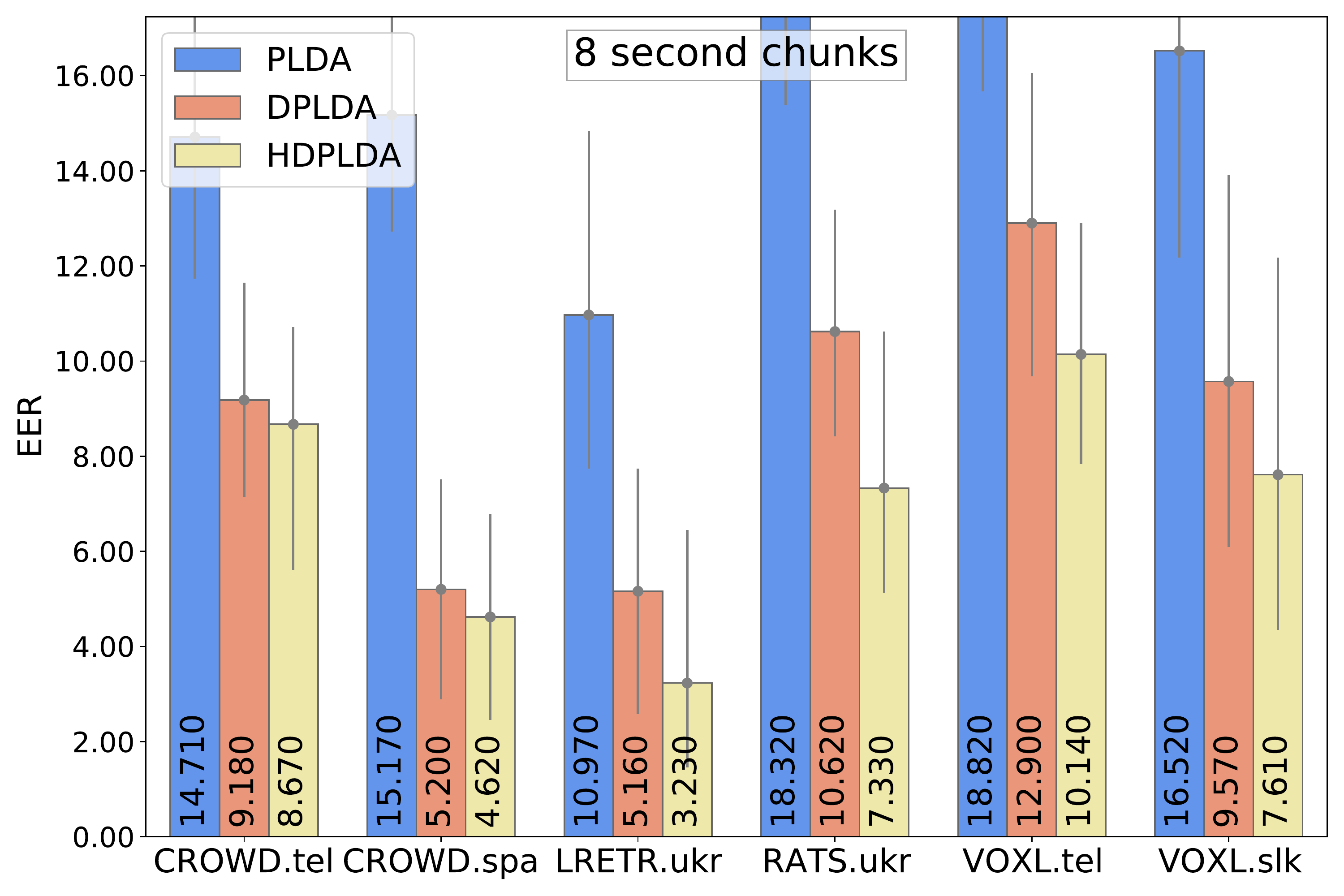}
\includegraphics[height=5cm]{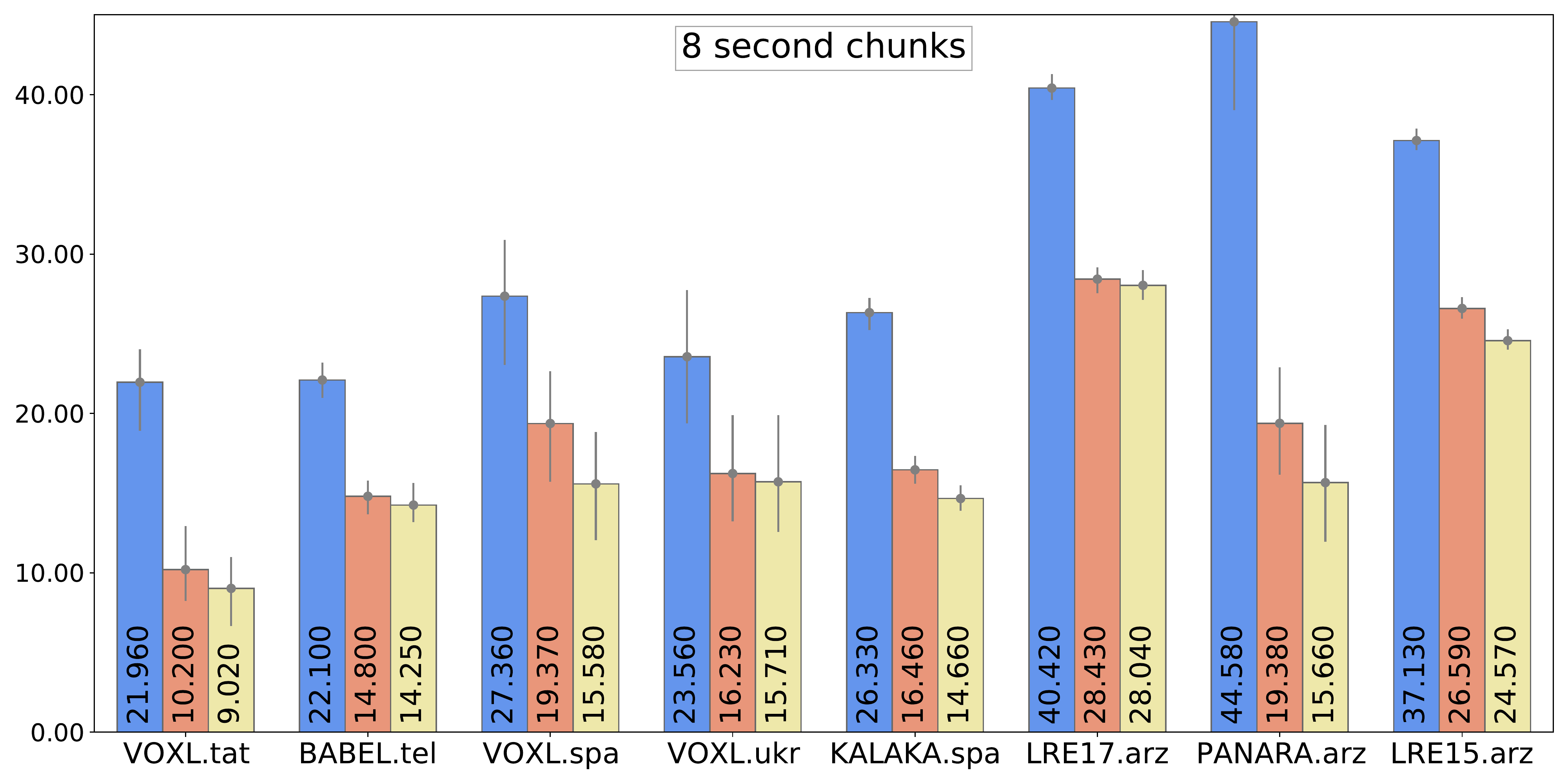}
\includegraphics[height=5cm]{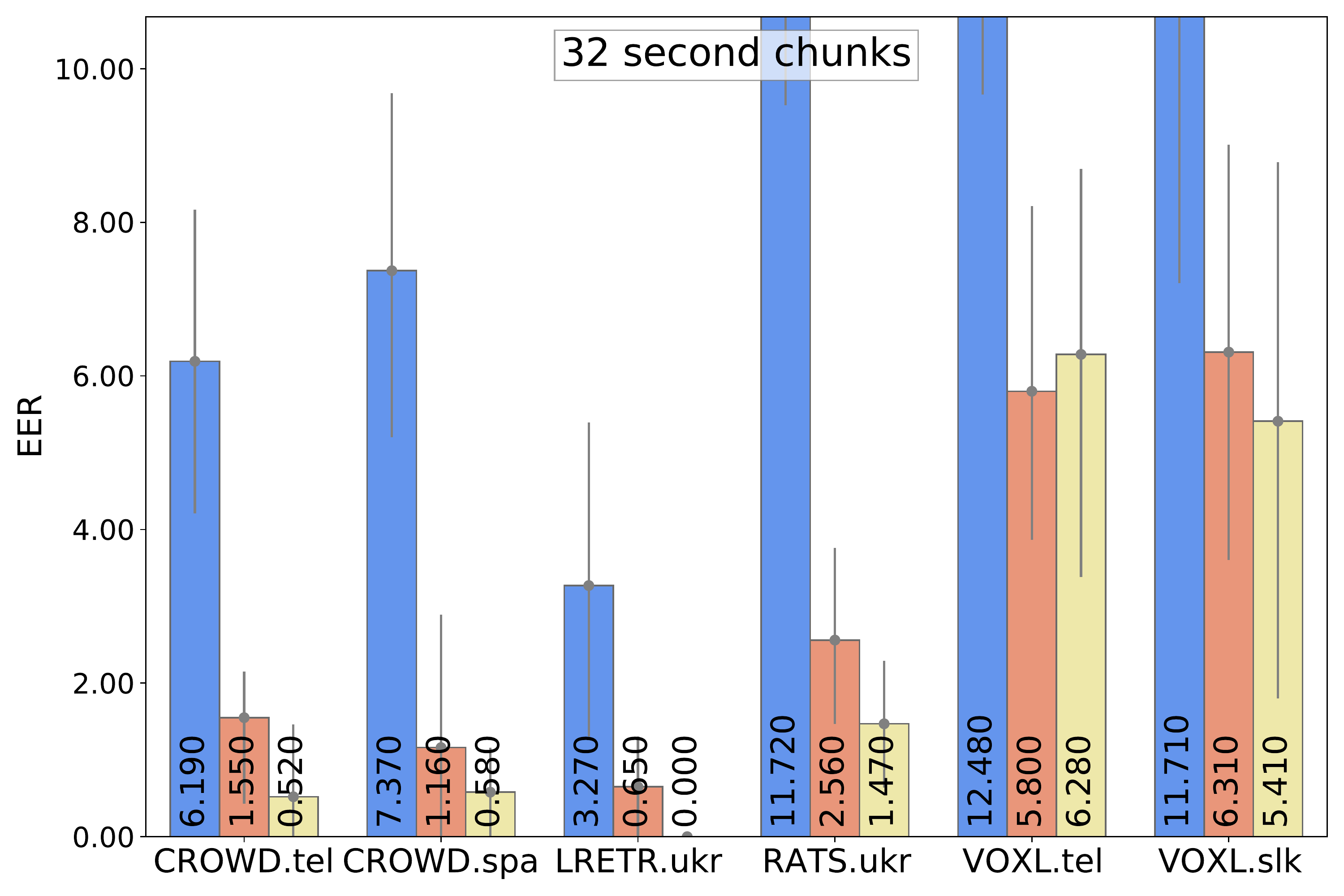}
\includegraphics[height=5cm]{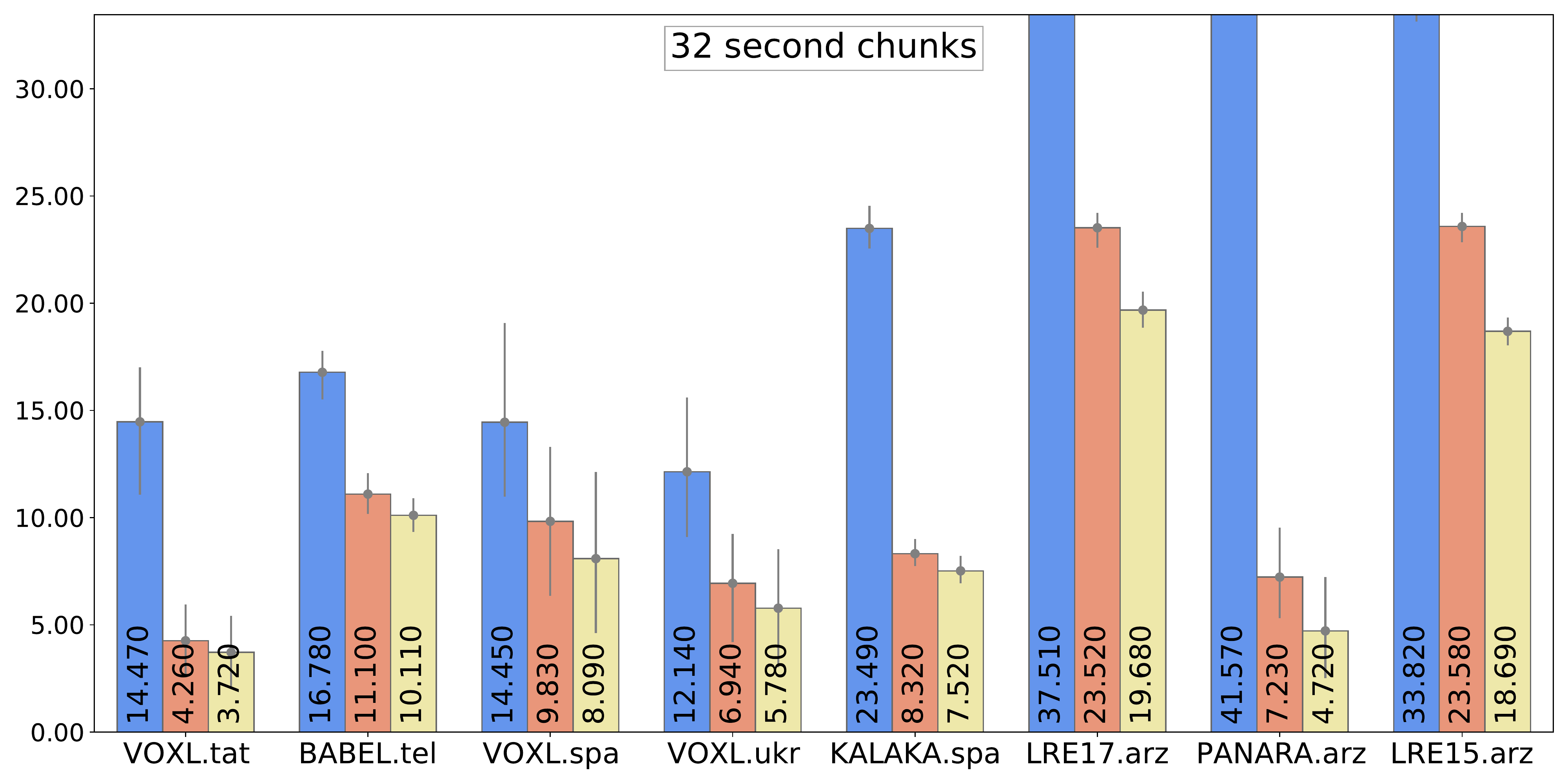}
\caption{Same as Figure \ref{fig:eval_eer} but with results computed over specific clusters using a threshold of 10. Sets are split and ordered as in Figure \ref{fig:eval_clust10}.}
\label{fig:eval_clust10_eer}
\end{figure*}

In this section, EER results are shown for all systems and datasets included in Figures \ref{fig:eval} and \ref{fig:eval_clust10}. In Figure \ref{fig:eval_eer} we can see that, when evaluating on all trials from each dataset, the EER values for all three systems stay under 9\% and 6\% for the 8- and 32-second chunks, respectively, with values under 1\% in several datasets. Large gains are observed when comparing DPLDA and HDPLDA with standard PLDA on some of the easier datasets. Figure \ref{fig:eval_clust10_eer} shows that evaluation by cluster is markedly harder, reaching EER values over 30\% in some datasets. In most cases, though, the proposed discriminative approaches lead to large gains with respect to PLDA.

\bibliographystyle{IEEEbib}
\footnotesize
\bibliography{all-short-no-crossrefs}

\end{document}